\documentclass{article}

  \PassOptionsToPackage{numbers, sort&compress}{natbib}
 \usepackage[preprint]{neurips_2026}

\usepackage[utf8]{inputenc} 
\usepackage[T1]{fontenc}    
\usepackage{hyperref}       
\usepackage{url}            
\usepackage{booktabs}       
\usepackage{amsfonts}       
\usepackage{nicefrac}       
\usepackage{microtype}      
\usepackage{xcolor}         
\usepackage{inconsolata}
\usepackage{graphicx}
\usepackage{amsmath}
\usepackage{multirow}
\usepackage{makecell}
\usepackage{xspace}
\usepackage[capitalize,noabbrev]{cleveref}
\usepackage{enumitem,kantlipsum}
\usepackage{arydshln}
\usepackage{subcaption}

\def\onedot{.\xspace}
\def\versus{\emph{vs}\onedot} 
\def\eg{\emph{e.g}\onedot} 
\def\ie{\emph{i.e}\onedot} 
 
\def\etc{\emph{etc}\onedot}

\usepackage{amsmath,amsfonts,bm}

\def\eqref#1{equation~\ref{#1}}

\def\1{\bm{1}}

\def\mC{{\bm{C}}}

\def\mH{{\bm{H}}}

\def\mP{{\bm{P}}}

\def\mR{{\bm{R}}}

\def\mU{{\bm{U}}}

\def\mW{{\bm{W}}}
\def\mX{{\bm{X}}}

\def\mLambda{{\bm{\Lambda}}}

\DeclareMathAlphabet{\mathsfit}{\encodingdefault}{\sfdefault}{m}{sl}
\SetMathAlphabet{\mathsfit}{bold}{\encodingdefault}{\sfdefault}{bx}{n}

\newcommand{\R}{\mathbb{R}}

\DeclareMathOperator{\clamp}{clamp}

\title{\emph{OffQ}: Taming Structured Outliers in LLM Quantization by Offsetting}

\author{%
  Haoqi Wang$^1$, Lorenz K. Mueller$^2$, Jiawei Zhuang$^2$, Mathieu Salzmann$^{13}$, Lukas Cavigelli$^2$\thanks{corresponding author} \\
  $^1$School of Computer and Communication Sciences, EPFL, Switzerland\\
  $^2$Huawei, Switzerland\\
  $^3$Swiss Data Science Center, ETHZ \& EPFL, Switzerland\\
  \texttt{\{haoqi.wang,mathieu.salzmann\}@epfl.ch}\\
  \texttt{\{lorenz.mueller,zhuangjiawei,lukas.cavigelli\}@huawei.com}
}

\begin{document}

\maketitle

\begin{abstract}
  Low-bit quantization has been widely adopted to accelerate the inference of large language models (LLMs) by significantly reducing computational cost and memory usage.
However, activation outliers pose a major challenge to effective quantization, often leading to notable performance degradation.
In this paper, we introduce OffQ, a method designed to mitigate activation outliers in low-bit quantization through a novel offsetting mechanism.
Specifically, OffQ first identifies a low-dimensional outlier subspace in the activations using a proposed top-1 PCA, and then concentrates high-magnitude activations into 1 channel via rotation.
OffQ then absorbs this concentrated outlier channel by converting its magnitude into a shared offset, thereby reducing the standard deviation of the activations.
This offsetting strategy enables effective W4A4KV4 quantization of LLMs using deployment-friendly uniform-grid and uniform-precision quantization.
Extensive experiments across diverse LLM architectures and benchmarks demonstrate that OffQ outperforms state-of-the-art baselines, consistently improving model accuracy while preserving low-bit efficiency.

\end{abstract}

\section{Introduction}\label{sec:introduction}

Quantization~\cite{krishnamoorthi2018quantizing,frantar2023optq,ashkboos2024quarot,liu2025spinquant,saxena2025resq} is a key technique for accelerating inference speed and reducing the memory footprint of large language models (LLMs).
Its importance is further amplified by the increasing demand to deploy LLMs in resource-constrained environments, such as edge devices and cost-sensitive cloud platforms.
By representing weights and activations with low-bit (\eg, 4-bit) integers rather than 16-bit floating-points, quantization significantly reduces model storage and improves computational efficiency.
As LLM inference is often memory-bound, reducing memory traffic directly translates into higher throughput, lower latency, and lower serving cost~\cite{MLSYS2024_5edb57c0,lin2025qserve}.

Despite this promise, robust 4-bit quantization of W4A4KV4 (Weight, Activation, and KV-cache all quantized in 4-bit) for LLMs remains challenging.
A key difficulty stems from the activation outliers: sparse but extremely large values in a few channels that dominate the dynamic range~\cite{bondarenko2021understanding,dettmers2022llmint8}.
These outliers interact poorly with quantization because they force a coarse quantization step size that is suboptimal for the majority of the activations, leading to large rounding errors.
A breakthrough in outlier mitigation is the use of rotation-based techniques, such as QuaRot~\cite{ashkboos2024quarot} and SpinQuant~\cite{liu2025spinquant}, which apply orthonormal transformations to redistribute the activation energy to Gaussian-like distributions, significantly improving quantization quality.
Yet, even equipped with the state-of-the-art rotation-based methods, the resulting activation distribution still retain large variance and is suboptimal for quantization~\cite{shao2026dartquant},
rendering W4A4KV4 quantization unreliable.

Several strategies have been explored to further improve quantization on top of rotation-based methods.
The first approach either improves the rotation learning or uses more expressive equivalent transformations (\eg, OSTQuant~\cite{hu2025ostquant}, AffineQuant~\cite{ma2024affinequant}, DuQuant~\cite{lin2024duquant}, FlatQuant~\cite{sun2025flatquant}, DartQuant~\cite{shao2026dartquant}, KurTail~\cite{akhondzadeh-etal-2025-kurtail}) to transform activations into a quantization-friendly form, at the cost of additional parameters, computational overhead, and carefully tuned custom kernels.
The second approach adopts mixed-precision schemes (\eg, SliM-LLM~\cite{pmlr-v267-huang25aa}, Atom~\cite{MLSYS2024_5edb57c0}, ResQ~\cite{saxena2025resq}), preserving higher precision for outlier-heavy channels while aggressively quantizing the rest.
However, it introduces heterogeneous execution, requiring extra kernels, casting, and memory movement, and often reducing the efficiency of streamlined low-bit computation.
The third approach uses non-uniform quantization (\eg, RCP~\cite{choi-etal-2025-rotate}, NestQuant~\cite{savkin2025nestquant}), which adapts quantization levels to the weight or activation distribution, but requires more complex kernels and lacks broad hardware support.

By contrast, here, we exploit recent progress in singular defects analysis~\cite{wang2025demystifying}, which unveils that the outliers in activation exhibit a pronounced \emph{low-dimensional structure}: the high-norm tokens in an LLM share the same direction across layers, token ids, and locations in a sequence.
Based on this observation, we introduce \emph{\textbf{OffQ}}, a simple, efficient, and effective approach to mitigating activation outliers for 4-bit quantization without resorting to mixed-precision components, non-uniform quantization levels, or compute and memory consuming backpropagation-based rotation learning.
The core idea is to eliminate the effect of outliers from the bulk of the activation distribution so that standard low-bit quantization can most effectively utilize its representational capacity.

Specifically,
OffQ first identifies the low-dimensional outlier subspace via a tailored top-1 PCA procedure and rotate the activations so that outliers concentrate on the least number of channels.
It then reorders the channels into groups so that each group is assigned with one outlier channel.
Finally, OffQ applies a specialized Hadamard rotation to convert the outlier energy into group-wise offsets, which will then be absorbed into the zero-point of standard low-bit quantization.
In essence, the offset within each group eliminates the corresponding outlier channel, yielding quantities with substantially reduced standard deviation that can be effectively quantized.
By suppressing outliers via offsets, OffQ enables accurate W4A4KV4 quantization while keeping all matrix multiplications in uniform 4-bit execution.

We evaluate OffQ on a range of LLM architectures and benchmarks, focusing on the challenging W4A4KV4 quantization setting.
Across these experiments, OffQ consistently improves the perplexity and accuracy compared to a range of state-of-the-art baselines, while preserving the simplicity and efficiency of end-to-end low-bit inference.
Overall, OffQ offers a practical mechanism for making 4-bit quantization more robust, strengthening the feasibility of low-cost LLM deployments.

\paragraph{Contributions}
\begin{enumerate}[leftmargin=0.5cm]
  \item We identify and address the structured activation outliers that limit practical low-bit quantization.
  \item We propose OffQ, a post-training quantization method that (i) rotates activations to concentrate outliers into a few channels, and (ii) absorbs them as group-wise offsets using Hadamard rotation, enabling W4A4KV4 quantization with uniform-precision matrix multiplication.
  \item Extensive experiments across multiple LLM architectures and benchmarks demonstrate that OffQ significantly improves 4-bit quantization accuracy while maintaining low-bit efficiency.
\end{enumerate}

\section{Related Work}\label{sec:related_work}

\paragraph{Quantization-Aware Training and Post-Training Quantization.}

Quantization in LLMs can be broadly categorized into Quantization-Aware Training (QAT)~\cite{jacob2018quantization, chen-etal-2025-efficientqat, liu-etal-2024-llm,ashkboos2025halo,onebit2024xu,2310.11453,du-etal-2024-bitdistiller,kim-etal-2022-understanding,oneill-dutta-2023-self} and Post-Training Quantization (PTQ)~\cite{arai2025quantization,kim2024squeezellm,wei2022outlier,dettmers2022llmint8,huang2024billm,li2025gptaq,vanbaalen-gptvq,vptq,xu2025crvq}.
QAT simulates quantization effects during training, allowing the model to adapt to low-precision representations.
However, the retraining involved can be computationally and memory-wise prohibitive.
In contrast, PTQ applies quantization to pre-trained models without further training, reducing its computational and memory requirements.
In this work, we focus on PTQ due to its efficiency and practicality for LLMs.

\paragraph{Weight-Only, Weight-Activation, and KV-Cache Quantization.}

In the context of PTQ for LLMs, weight-only quantization~\cite{frantar2023optq,muller2025sinq,dettmers2024spqr,pmlr-v235-kim24f,tseng2024qtip,zhang2024afpq,lin2023awq,hu2025mlwq,guo2024lqlora,zhang2025leanquant,lee2023flexround,zhang2024magr,arai2026quantization,chee2024quip2bitquantizationlarge,tseng2024quip,pmlr-v267-huang25aa,adepuFQIcml24} focuses on quantizing the model weights while keeping activations in higher precision.
It reduces memory traffic during decoding phase of the inference, but requires the costly full-precision matrix multiplication and cannot fully leverage the emerging efficient low-bit hardware.
By contrast, weight-activation quantization, which quantizes both weights and activations, not only reduces memory but also speeds up computation in compute-bound scenarios such as the prefill phase and batched inference~\cite{QUIK,kurtic-etal-2025-give}, as arithmetics can be performed in low-bit using specialized hardware such as Tensor Core~\cite{lin2023awq,ashkboos2024quarot,MLSYS2024_5edb57c0,Thakkar_CUTLASS_2023}.
KV-cache quantization~\cite{su2025accurate,cai2025pyramidkv,yang-etal-2025-xquant,he2024zipcache,hooper2024kvquant,liu2024kivi,son2025nsnquant,zandieh2026turboquant} reduces the memory bandwidth overhead of autoregressive decoding by caching key/value in low-bit.
It can be applied in conjunction with weight-only and weight-activation quantization.
In this work, we focus on the challenging but rewarding W4A4KV4 setting: applying weight-activation quantization together with KV-cache quantization.

\paragraph{Equivalent Transforms in Weight-Activation Quantization.}

A central challenge in weight-activation quantization is the presence of outliers in activations.
While techniques such as mixed-precision~\cite{dettmers2022llmint8,QUIK,pmlr-v267-huang25aa,MLSYS2024_5edb57c0} are sometimes used, existing solutions mainly rely on the idea of equivalent transform:
applying a linear transformation to the activations before quantization and applying the inverse transformation after dequantization, which does not change the model's representational capacity but can redistribute the activation energy to be less spiky and more quantization-friendly.
Prior work mainly uses the following types of equivalent transforms:
\begin{enumerate}[leftmargin=0.7cm, label=(\roman*)]
    \item  \emph{Permutation.}
          RPTQ~\cite{yuan2023rptq} permutes channels so that similar value ranges are clustered into groups and apply fine-grained quantization.

    \item \emph{Scaling and Shifting}, either learnable or heuristic-based, are used in SmoothQuant~\cite{xiao2023smoothquant}, Outlier Suppressions~\cite{wei2022outlier,wei-etal-2023-outlier}, OmniQuant~\cite{shao2024omniquant}, and ABQ-LLM~\cite{zeng2025abq}.

    \item \emph{Rotation.}
          Rotation as an equivalent transform was initially used in SliceGPT~\cite{ashkboos2024slicegpt} for model trimming, and later widely adopted in the quantization literature due to its effectiveness in outlier suppression.
          QuaRot~\cite{ashkboos2024quarot} uses random Hadamard rotation;
          SpinQuant~\cite{liu2025spinquant} trains rotation by Cayley SGD;
          DartQuant~\cite{shao2026dartquant} optimizes the rotation using Whip loss;
          DFRot~\cite{xiang2025dfrot} learns rotation with a weighted loss;
          KurTail~\cite{akhondzadeh-etal-2025-kurtail} learns rotation by a Kurtosis loss;
          DuQuant~\cite{lin2024duquant} combines two rotations with permutation;
          ResQ~\cite{saxena2025resq} uses rotation and computes \(\nicefrac{1}{8}\) of the channels in higher precision;
          Both NestQuant~\cite{savkin2025nestquant} and RCP~\cite{choi-etal-2025-rotate} use rotation combined with non-uniform grid.

    \item \emph{Affine Transform.}
          AffineQuant~\cite{ma2024affinequant} and FlatQuant~\cite{sun2025flatquant} learn affine transforms using an MSE loss;
          QLLM~\cite{liu2024qllm} reallocates outlier magnitudes to other channels;
          OSTQuant~\cite{hu2025ostquant} and QServe~\cite{lin2025qserve} combine rotation and scaling.
\end{enumerate}

Many of the aforementioned equivalent transform-based methods also combine mixed-precision~\cite{liu2024qllm,saxena2025resq}, and non-uniform grid~\cite{savkin2025nestquant,choi-etal-2025-rotate} to further enhance the quantization performance.
In this work, we introduce a new equivalent transform:

\begin{enumerate}[leftmargin=0.7cm, label=(\roman*), start=5]
    \item \emph{Outlier2offset}, that converts outliers into offsets.
          It is complementary to the existing techniques (backpropagation-based rotation optimization, mixed-precision, non-uniform quantization grid, \etc) and can be seamlessly combined with them for further improvement.
          Nevertheless, we demonstrate that our OffQ can achieve strong performance with the standard uniform-grid and uniform-precision setting, and leave the exploration of advanced combinations for future work.
\end{enumerate}

\section{Method}\label{sec:method}

Quantization maps high-precision floating-point values to lower-bit discrete representations, reducing memory footprint and accelerating inference.
We adopt a \emph{uniform quantizer}~\cite{krishnamoorthi2018quantizing},  which transforms a floating-point value \(x\) into a \(b\)-bit integer \(x_q\) via:
\begin{equation}
  x_q = \clamp\left(\left\lfloor\frac{x}{s}\right\rceil+z, 0, 2^{b}-1\right),
\end{equation}
where \(\lfloor\,\cdot\,\rceil\) denotes the rounding operator, \(s\) is the scale factor defining the quantization step size, and \(z\) is the zero-point, an integer offset that enables \emph{asymmetric quantization}.
Omitting \(z\) gives \emph{symmetric quantization}.
The granularity of quantization parameters can be adjusted.
\emph{Per-tensor quantization} uses a single pair \((s, z)\) for an entire weight or activation tensor;
while memory-efficient, it suffers from quantization error in the presence of outliers.
\emph{Per-channel quantization} assigns separate parameters to each output channel, better capturing channel-wise distributional variances.
\emph{Per-group quantization} further partitions each channel into groups with dedicated parameters, offering finer-grained representational fidelity at the cost of a marginal increase in metadata storage.

\subsection{Concentrate Structured Outliers via Top-1 PCA}

\begin{figure}[t]
  \begin{subfigure}[b]{0.245\textwidth}
    \includegraphics[width=\textwidth]{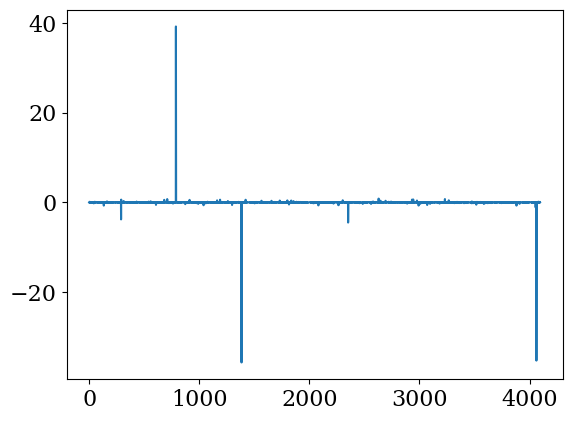}
    \caption{Token \(\mX \in \R^{1\times4096}\)}
    \label{fig:x}
  \end{subfigure}
  \hfill
  \begin{subfigure}[b]{0.245\textwidth}
    \includegraphics[width=\textwidth]{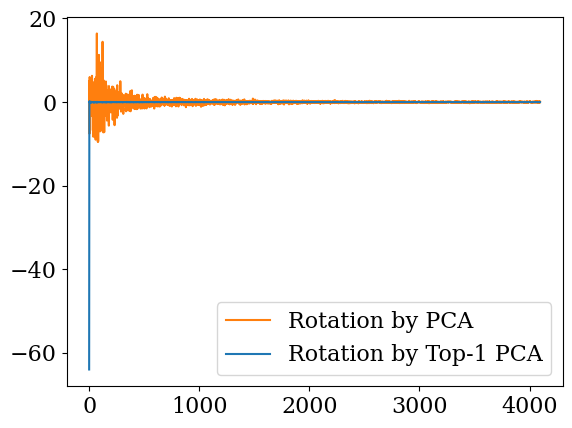}
    \caption{Concentrated \(\mX_\text{rot}\)}
    \label{fig:top1}
  \end{subfigure}
  \hfill
  \begin{subfigure}[b]{0.245\textwidth}
    \includegraphics[width=\textwidth]{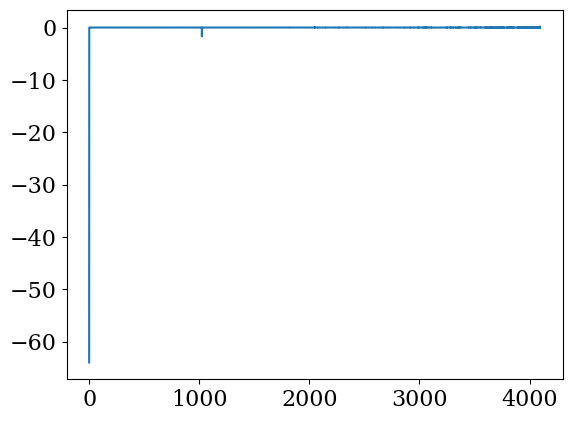}
    \caption{Permute to 4 groups}
    \label{fig:groups}
  \end{subfigure}
  \hfill
  \begin{subfigure}[b]{0.245\textwidth}
    \includegraphics[width=\textwidth]{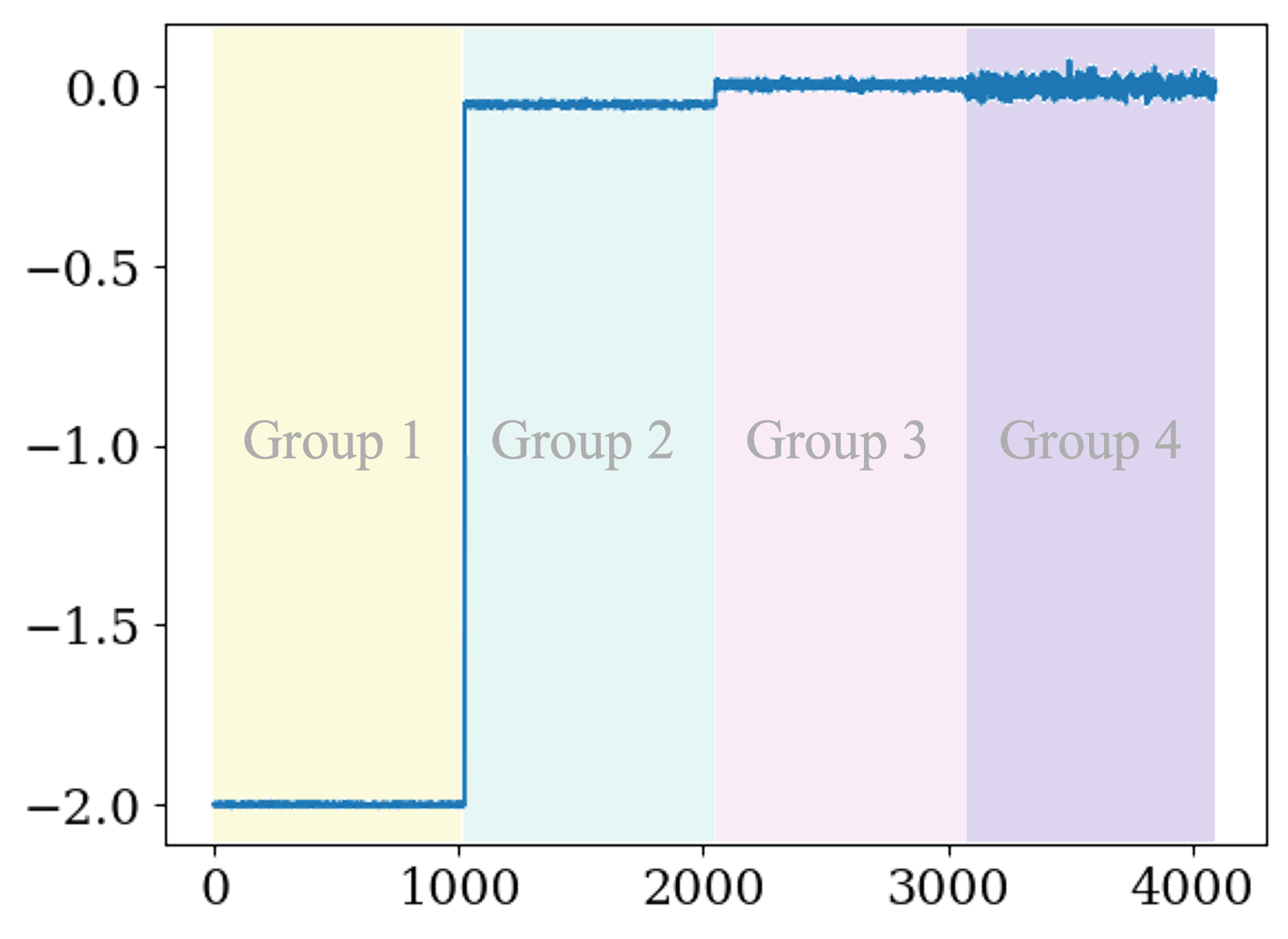}
    \caption{Group-wise offset \(\mX_\text{had}\)}
    \label{fig:offsetting}
  \end{subfigure}
  \caption{
    Visualization of the outlier concentration of the top-1 PCA and the subsequent offsetting technique.
    We plot the activation values of an outlier token in Llama 3-8B before and after each step, where the \(x\)-axis is the channel index and the \(y\)-axis is the activation value.
    The top-1 PCA effectively concentrates the outlier into the first channel, while the regular PCA fails to do so due to interference from normal tokens.
    The permutation and group-wise offsetting then redistributes the outlier value across all channels, creating constant offsets that can be absorbed during quantization.
  }\label{fig:pca}
\end{figure}

The emergence of structured outliers in LLMs has been widely documented in the quantization literature~\cite{bondarenko2021understanding,dettmers2022llmint8} and in studies of massive activations~\cite{sun2024massive}.
These outliers are characterized by their sparsity and extremely large magnitudes.
A key insight from recent work on singular defects in LLMs~\cite{wang2025demystifying} is that such activation outliers exhibit a pronounced \emph{low-dimensional structure}.
Specifically, the outlier directions across the transformer layer boundary are dominated by a 1-dimensional subspace and the subspace is shared across different layers, token IDs, and locations in a sequence.
This observation suggests that the intrinsic dimensionality of the outlier subspace is substantially smaller than the number of affected channels.
Consequently, as illustrated in \cref{fig:pca}, concentrating outliers into fewer channels can enable more targeted suppression techniques.

\paragraph{Top-1 PCA.}
To extract the low-dimensional outlier subspace, we propose a \emph{top-1 PCA} procedure.
Given a small calibration dataset of \(N\) sequences each of length \(L\), we collect input activations before each linear layer with hidden dimension \(D\), forming an \(N\times L\times D\) tensor.
To focus on outlier tokens and suppress interference from normal tokens, we select the top-1 token per sequence with the largest \(L^\infty\) norm, yielding a matrix \(\mX\in\mathbb{R}^{N\times D}\).
Assuming zero-mean activations, we compute the covariance matrix \(\mC\) and perform eigendecomposition:
\begin{equation}
  \mC = \left(\mX^\top\! \mX\right)\!/ N = \mU\! \mLambda \mU^\top,
\end{equation}
where the columns of $\mU$ are the eigenvectors sorted by their corresponding eigenvalues in $\mLambda$ in descending order.
The leading eigenvector (first column of $\mU$) captures the direction of maximal outlier variance.

\paragraph{Concentrate Outliers.}
To align the outlier directions with individual channels, we rotate the activations \(\mX\) with the eigenvector matrix \(\mU\),
\begin{equation}
  \mX_{\text{rot}} = \mX \mU.
\end{equation}
The first channel of $\mX_{\text{rot}}$ corresponds to the projection onto the leading eigenvector, thereby concentrating the dominant outlier into a single channel.
More generally, the leading \(G\) channels capture the dominant outlier directions of the original activations, while the remaining channels are largely outlier-free and thus amenable to low-bit quantization.

\paragraph{Why the \(L^\infty\) Norm.}
We assume a pre-normalization architecture with RMSNorm, consistent with most contemporary LLMs.
Following QuaRot~\cite{ashkboos2024quarot} and SpinQuant~\cite{liu2025spinquant}, we fuse the scaling parameters of RMSNorm into the adjacent weight matrices, so that the input activations to the linear layers of \(\mW_Q, \mW_K, \mW_V\) in the self-attention module and \(\mW_U, \mW_G\) in the feedforward module, lie on a scaled \(L^2\) sphere.
On a sphere,
tokens with a few large-magnitude entries and many small entries exhibit a large \(L^\infty\) norm.
For example, on the 3-dimensional unit sphere, the spiky token \([1, 0, 0]\) has a large \(L^\infty\) norm of \(1\), while the flat token \([0.58, 0.58, 0.58]\) has a smaller \(L^\infty\) norm of \(0.58\).
The \(L^\infty\) norm therefore serves as a proxy for identifying the extreme outlier tokens in each sequence.

\paragraph{Why Top-1 Selection.}
While prior work has applied PCA to identify the outlier subspace for quantization~\cite{saxena2025resq}, our top-1 PCA is more effective at concentrating outliers by
focusing exclusively on the most extreme tokens, which are the primary drivers of quantization error.
Because activations are \(L^2\)-normalized, the covariance matrix computed over all tokens is distracted by the large number of normal tokens, causing the leading eigenvector to drift from the true outlier direction.
\cref{fig:top1} validates this intuition, showing that top-1 PCA yields better outlier concentration than standard PCA.

\vspace{-0.25em}
\paragraph{Types of Structured Outliers.}
We identify two primary types of structured outliers in LLMs.
The first type arises from singular defects~\cite{sun2024massive,wang2025demystifying}, appearing at the boundary between the self-attention and feedforward modules.
The second type occurs in LLMs that incorporate bias terms in self-attention layers (e.g., Qwen~2.5~\cite{qwen2025qwen25technicalreport}), where these biases skew the activation distribution and introduce outliers into the KV-cache.
Both types are effectively captured by the Top-1 PCA.

\vspace{-0.25em}
\subsection{Absorbing Outliers by Offsetting}

\vspace{-0.25em}
Having concentrated the prominent outlier into the first channel of the rotated activations \(\mX_{\text{rot}}\), we next introduce our offsetting technique to absorb this outlier prior to quantization.

\vspace{-0.25em}
\paragraph{Offsetting Technique.}
We construct a partially random Hadamard matrix \(\mH \in \mathbb{R}^{D \times D}\), an orthogonal matrix with \(\pm1\) entries, and whose first row is constrained to be all ones,
\ie, $\mH_{1,:} = [1, 1, \ldots, 1]$.
We then apply the Hadamard rotation \(\mH\) to the outlier-concentrated activations \(\mX_\text{rot}\),
\begin{equation}
  \mX_{\text{had}} = \mX_{\text{rot}} \mH/\sqrt{D}.
\end{equation}
This Hadamard rotation redistributes the value of the first channel \emph{uniformly} across all channels, creating a constant offset of \(\nicefrac{1}{\sqrt{D}}\) times  the original outlier value into every channel.
To illustrate this, consider a 2-dimensional token $\mX_{\text{rot}} = [x_1, x_2] \in \mathbb{R}^{1\times2}$ with an outlier in the first channel $x_1$.
Applying the Hadamard rotation with $\mH = \begin{bmatrix} 1 & 1 \\ 1 & -1 \end{bmatrix}$ gives:
\begin{align}
  \mX_{\text{had}} = \mX_{\text{rot}} \mH/\sqrt{2} & = \left[\frac{x_1 + x_2}{\sqrt{2}}, \frac{x_1 - x_2}{\sqrt{2}}\right]
  =\underbrace{\frac{x_1}{\sqrt{2}}}_{\text{offset}} + \underbrace{\left[\frac{x_2}{\sqrt{2}}, -\frac{x_2}{\sqrt{2}}\right]}_{\text{without outlier } x_1}.
\end{align}
The outlier $x_1$ is spread across all channels as a constant offset \(\nicefrac{x_1}{\sqrt{2}}\), which can be absorbed into the zero-point under asymmetric quantization\footnote{Asymmetric quantization needs not maintain the value 0 into the quantization range~\cite{krishnamoorthi2018quantizing}, since 0 is shifted by the offset.}.
The remaining activations exhibit a flatter distribution and can be quantized effectively.
Note that we use the Hadamard rotation for its simplicity and efficiency, but the offsetting technique is compatible with any orthogonal matrix whose first row contains only one unique value.
Future work may involve extending optimization frameworks like
DartQuant~\cite{ashkboos2024quarot}, DFRot~\cite{xiang2025dfrot}, or KurTail~\cite{akhondzadeh-etal-2025-kurtail}
to learn constrained rotations that further flatten the non-outlier channels.

\paragraph{Group-Wise Offsetting.}

To absorb multiple outlier channels, we extend the offsetting technique to a group-wise setting.
We partition the \(D\) channels of $\mX_{\text{rot}}$ into $G$ groups of size $\nicefrac{D}{G}$, and reorder the channels so that the first channel of the \(g\)-th group corresponds to the \(g\)-th largest outlier channel.
The offsetting technique is then applied independently within each group, so that the top-\(G\) outlier channels are each absorbed as a zero-point under per-group asymmetric quantization.
The remaining non-outlier channels are assigned to groups in sorted order similar to~\cite{yuan2023rptq}, which we found empirically to outperform the zigzag assignment of~\cite{lin2024duquant}.

\subsection{Quantizing an LLM with Activation Offsetting}

\begin{figure}[t]
  \centering
  \includegraphics[width=0.99\textwidth]{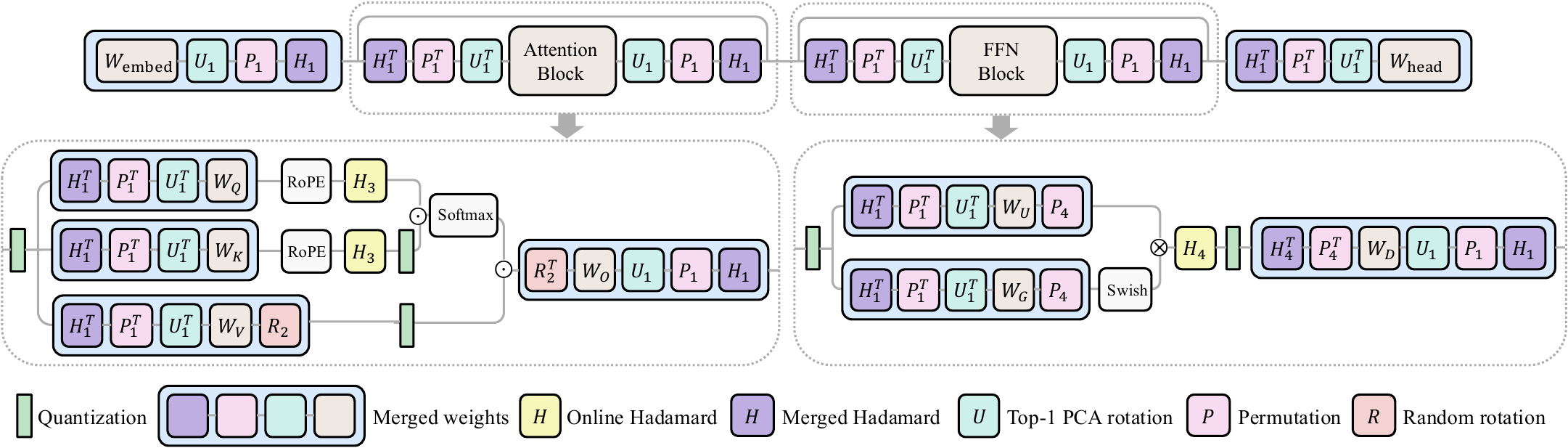}
  \caption{
    The overall quantization pipeline of OffQ.
    \(U_1\) is the shared rotation from top-1 PCA on attention and feedforward inputs, \(P_1\) and \(P_4\) are the permutation matrices for grouping, \(R_2\) is a per-head random rotation, \(H_1\) is the group-wise Hadamard rotation for offsetting, and \(H_3, H_4\) with yellow background are the online Hadamard rotations.
    Most rotation matrices, except for the three online Hadamard rotations, are fused into the weight matrices to reduce computational overhead.
  }\label{fig:pipeline}
  \vspace{-0.5em}
\end{figure}

Building upon our insights into the structured nature of outliers and the offsetting technique, we introduce \emph{OffQ}, a novel quantization method that effectively mitigates outlier-induced quantization error.
The overall pipeline, illustrated in Figure~\ref{fig:pipeline}, enhances the architecture of the rotation-based method SpinQuant~\cite{liu2025spinquant} by incorporating our activation offsetting technique.

\paragraph{Step 1. Collect Activation Statistics.}
We begin by fusing the RMSNorm scaling parameters into the adjacent weight matrices.
Activation statistics are then collected at the following positions:
(1)~\emph{Attention and feedforward inputs.}
Input activations to \(\mW_Q, \mW_K, \mW_V, \mW_U, \mW_G\) are stacked together over all layers.
The rotation \(\mU_1\) is computed via Top-1 PCA on these activations and shared among layers.
(2)~\emph{Key activations.}
If the LLM contains bias-induced outliers in self-attention, we additionally collect the output activations of \(\mW_K\) after RoPE encoding.
The corresponding rotation \(\mU_3\) is computed via per-layer, per-head Top-1 PCA.
(3)~\emph{Down-projection inputs.}
We collect the per-channel absolute maxima of input activations to \(\mW_{\text{D}}\) separately per layer.

\paragraph{Step 2. Apply Rotation, Grouping, and Offsetting.}
These operations can be represented by orthonormal matrices and are fused into the weight matrices when possible to reduce the overhead.
\begin{enumerate}[leftmargin=0.64cm, label=(\arabic*), itemsep=2pt, topsep=0pt]
  \item \emph{Attention and feedforward inputs.}
        We right-multiply \(\mW_O\) and \(\mW_D\) by \(\mU_1\), apply the permutation \(\mP_1\) to reorder channels into groups, and apply the group-wise Hadamard rotation \(\mH_1\) for offsetting.
        The inverse transformation \(\mH_1^\top \mP_1^\top \mU_1^\top\) is left-multiplied onto \(\mW_Q, \mW_K, \mW_V, \mW_U, \mW_G\) to preserve the original output space.
        The input embeddings are also rotated by \(\mU_1 \mP_1 \mH_1\), with the inverse applied to the left of the output head \(\mW_\text{head}\).
  \item  \emph{Value cache.}
        A per-head random rotation \(\mR_2\) is fused into the right side of \(\mW_V\) and the left side of \(\mW_O\), enabling quantization of the value cache.
  \item \emph{Key cache.}
        If no bias-induced outliers are present, we apply the per-head online Hadamard rotation \(\mH_3\) to the RoPE-encoded outputs of \(\mW_Q\) and \(\mW_K\).
        Otherwise, \(\mH_3\) is replaced by \(\mU_3 \mP_3 \mH_3\), which prepends a per-layer, per-head top-1 PCA rotation \(\mU_3\) and a permutation \(\mP_3\).
  \item \emph{Down-projection inputs.}
        Because rotation cannot penetrate the element-wise multiplication preceding \(\mW_D\), top-1 PCA is not applicable here.
        Instead, outlier channels are identified via per-channel absolute maxima.
        Channels are permuted into groups by \(\mP_4\), and then offset by online group-wise Hadamard rotation \(\mH_4\).
        The permutation \(\mP_4\) is fused into the right side of \(\mW_U, \mW_G\), and the inverses \(\mH_4^\top \mP_4^\top\) are fused into the left side of \(\mW_D\).
\end{enumerate}

\paragraph{Step 3. Quantize Weights.}

Following the practice of rotation-based quantization methods such as QuaRot~\cite{ashkboos2024quarot}, SpinQuant~\cite{liu2025spinquant}, DartQuant~\cite{shao2026dartquant}, Atom~\cite{MLSYS2024_5edb57c0}, and KurTail~\cite{akhondzadeh-etal-2025-kurtail}, we apply  GPTQ~\cite{frantar2023optq} to quantize the weights in 4 bits after merging of transforms into the weight matrices.
More advanced weight quantizers such as GPTAQ~\cite{li2025gptaq}, QEP~\cite{arai2025quantization}, LeanQuant~\cite{zhang2025leanquant}, Qronos~\cite{zhang2026qronos}, and ResComp~\cite{li2026rethinking} are compatible with our pipeline, but we use GPTQ for all methods to ensure a fair comparison.

\subsection{Discussion}

\paragraph{Generality on Data Formats.}
Micro-scaling formats MXFP4 and NVFP4 employ block-wise shared scale factors and are increasingly supported by recent hardware.
Applying the offsetting technique to these formats requires extending them to support asymmetric quantization, which introduces a zero-point parameter per block.
With a suitable block size, the overhead of zero-points remains negligible while still effectively absorbing outliers.
We leave this extension to future work.

\vspace{-0.5em}
\paragraph{Application to ViTs.}
Vision Transformers (ViTs) are also known to exhibit structured outliers~\cite{darcet2024vision,wang2024sinder}, making OffQ a natural candidate for extension. We leave this exploration to future work.

\paragraph{Efficiency of Per-Group Activation Quantization.}
Per-group quantization is widely used in weight quantization~\cite{frantar2023optq}, and recent work including Atom~\cite{MLSYS2024_5edb57c0} and QServe~\cite{lin2025qserve}
have applied per-group quantization to activations with custom kernel implementations.
We expect advances in hardware support and kernel design to further reduce the computational overhead of group quantization.
In terms of memory, the zero-point storage overhead per group is negligible relative to the savings from 4-bit quantization.
For example, with a group size of 128 and scale/zero-point stored in 16 bits each, the effective bit-width per activation is \(\nicefrac{(4\times128+16\times2)}{128} = 4.25\) bits, which is still a substantial reduction from the original 16 bits.

\section{Experiments}\label{sec:experiment}

\begin{table}[t]
    \caption{
        Compare the perplexity (PPL \(\downarrow\)) on WikiText~\cite{merity2017pointer} and the average 0-shot accuracy (0-shot\(^8\)~\(\uparrow\)) on ARC-e/ARC-c~\cite{clark2018thinksolvedquestionanswering}, BoolQ~\cite{clark-etal-2019-boolq}, HellaSwag~\cite{zellers2019hellaswag}, OpenBookQA~\cite{mihaylov-etal-2018-suit}, PIQA~\cite{Bisk2020}, SIQA~\cite{sap2019social}, and WinoGrande~\cite{sakaguchi2019winogrande} under W4A4KV4 quantization.
        Results for 16-bit, GPTQ~\cite{frantar2023optq}, QUIK~\cite{QUIK}, QuaRot~\cite{ashkboos2024quarot}, SpinQuant~\cite{liu2025spinquant}, and ResQ~\cite{saxena2025resq} are referenced from \cite{saxena2025resq};
        and results for DFRot~\cite{xiang2025dfrot}, KurTail~\cite{akhondzadeh-etal-2025-kurtail}, and OSTQuant~\cite{hu2025ostquant} are referenced from their respective papers.
        Missing results are marked with ``--''.
        OffQ achieves the best performance.
        Detailed results are provided in the Appendix.
    }\label{tab:result}
    \centering
    \begin{tabular}{l@{\hspace{5pt}}l@{\hspace{0pt}}c@{\hspace{0.pt}}c@{\hspace{0.pt}}c@{\hspace{0.pt}}c@{\hspace{3.pt}}c@{\hspace{0.pt}}c@{\hspace{3.pt}}c@{\hspace{0.pt}}c@{\hspace{3.pt}}c@{\hspace{0.pt}}c@{\hspace{3.pt}}c@{\hspace{0.pt}}c@{\hspace{1.pt}}c@{\hspace{0.pt}}ccccc}
        \toprule
                                                  & \multirow{2.5}{*}{Method} & \multicolumn{2}{c}{3-8B} & \multicolumn{2}{c}{3-70B} & \multicolumn{2}{c}{3.2-1B} & \multicolumn{2}{c}{3.2-3B} & \multicolumn{2}{c}{2-7B} & \multicolumn{2}{c}{2-13B}                                                                                                           \\
        \cmidrule(lr){3-4}\cmidrule(lr){5-6}\cmidrule(lr){7-8}\cmidrule(lr){9-10}\cmidrule(lr){11-12}\cmidrule(l){13-14}
                                                  &                           & PPL                      & 0-shot\(^8\)              & PPL                        & 0-shot\(^8\)               & PPL                      & 0-shot\(^8\)              & PPL             & 0-shot\(^8\)   & PPL             & 0-shot\(^8\)   & PPL              & 0-shot\(^8\)   \\
        \midrule
        \multirow{11}{*}{\rotatebox{90}{Llama}}   & 16-bit                    & ~~6.1~~                  & 67.09                     & ~~ 2.9                     & 73.09                      & ~~9.8~~                  & 54.86                     & ~~7.8~~         & 62.73          & ~~5.5~~         & 64.15          & ~~4.9~~          & 66.45          \\
        \cdashline{2-15}\noalign{\vskip 3.5pt}
                                                  & GPTQ                      & 166.3                    & 39.79                     & 11655                      & 34.90                      & 108.9                    & 37.98                     & 178.3           & 40.34          & 9600            & 38.89          & 3120             & 35.83          \\
                                                  & QUIK                      & 14.2~~                   & 51.60                     & ~~8.0~~                    & 58.15                      & 21.8~~                   & 44.30                     & 15.8~~          & 48.74          & ~~7.5~~         & 56.99          & ~~6.8~~          & 60.21          \\
                                                  & QuaRot                    & ~~7.8~~                  & 62.10                     & ~~5.7~~                    & 67.56                      & 14.3~~                   & 49.01                     & 10.1~~          & 56.06          & ~~6.1~~         & 60.75          & ~~5.4~~          & 63.80          \\
                                                  & SpinQuant                 & ~~7.4~~                  & 63.76                     & ~~6.2~~                    & 65.68                      & 13.6~~                   & 48.78                     & ~~9.2~~         & 57.89          & ~~6.0~~         & 60.98          & ~~5.2~~          & 64.81          \\
                                                  & DFRot                     & ~~7.91                   & 62.35                     & ~~5.03                     & 68.98                      & --                       & --                        & --              & --             & ~~6.25          & 60.97          & ~~5.43           & 63.83          \\
                                                  & KurTail                   & ~~7.2~~                  & 64.63                     & ~~4.2~~                    & 70.69                      & 12.9~~                   & 50.11                     & ~~9.0~~         & 59.04          & ~~5.9~~         & 61.31          & ~~5.2~~          & 65.18          \\
                                                  & OSTQuant                  & ~~7.29                   & 64.70                     & ~~4.01                     & \textbf{71.16}             & --                       & --                        & --              & --             & ~~5.91          & \textbf{62.11} & ~~5.25           & 64.19          \\
                                                  & ResQ                      & ~~7.1~~                  & 63.91                     & ~~4.1~~                    & 71.14                      & 12.4~~                   & 50.11                     & ~~8.8~~         & 58.99          & ~~5.8~~         & 61.95          & ~~\textbf{5.1}~~ & \textbf{65.25} \\
                                                  & \textbf{OffQ}             & ~~\textbf{6.98}          & \textbf{65.49}            & ~~\textbf{3.88}            & 70.63                      & \textbf{12.32}           & \textbf{50.91}            & ~~\textbf{8.78} & \textbf{60.80} & ~~\textbf{5.77} & 61.99          & ~~5.11           & \textbf{65.25} \\
        \midrule\addlinespace[-0.1ex]
        \midrule
                                                  & \multirow{2.5}{*}{Method} & \multicolumn{2}{c}{1.5B} & \multicolumn{2}{c}{3B}    & \multicolumn{2}{c}{7B}     & \multicolumn{2}{c}{14B}    & \multicolumn{2}{c}{32B}  & \multicolumn{2}{c}{72B}                                                                                                             \\
        \cmidrule(lr){3-4}\cmidrule(lr){5-6}\cmidrule(lr){7-8}\cmidrule(lr){9-10}\cmidrule(lr){11-12}\cmidrule(l){13-14}
                                                  &                           & PPL                      & 0-shot\(^8\)              & PPL                        & 0-shot\(^8\)               & PPL                      & 0-shot\(^8\)              & PPL             & 0-shot\(^8\)   & PPL             & 0-shot\(^8\)   & PPL              & 0-shot\(^8\)   \\
        \midrule
        \multirow{7}{*}{\rotatebox{90}{Qwen 2.5}} & 16-bit                    & ~~9.3~~                  & 60.85                     & ~~8.0~~                    & 63.81                      & ~~6.8~~                  & 68.45                     & ~~5.3~~         & 70.61          & ~~5.0~~         & 70.44          & ~~3.9~~          & 73.41          \\
        \cdashline{2-15}\noalign{\vskip 3.5pt}
                                                  & GPTQ                      & 25770                    & 35.21                     & 9978                       & 35.10                      & 13594                    & 34.85                     & 5100            & 36.93          & 3891            & 38.53          & 37967            & 34.54          \\
                                                  & QUIK                      & 6614                     & 35.81                     & 15.5~~                     & 51.19                      & 260.3                    & 41.48                     & 10.5~~          & 57.66          & ~~9.6~~         & 59.08          & 8.3              & 61.90          \\
                                                  & QuaRot                    & 6600                     & 38.33                     & 68.8~~                     & 47.76                      & 4036                     & 38.36                     & ~~6.8~~         & 67.14          & ~~6.1~~         & 67.90          & 4.9              & 70.28          \\
                                                  & ResQ                      & 12.5~~                   & 55.26                     & ~~9.0~~                    & 61.13                      & ~~8.2~~                  & 65.29                     & ~~6.2~~         & 69.16          & ~~5.6~~         & 69.55          & ~~4.6~~          & 71.98          \\
                                                  & \textbf{OffQ}             & \textbf{11.35}           & \textbf{57.53}            & ~~\textbf{8.98}            & \textbf{61.47}             & ~~\textbf{7.66}          & \textbf{66.16}            & ~~\textbf{6.07} & \textbf{69.20} & ~~\textbf{5.52} & \textbf{69.59} & ~~\textbf{4.29}  & \textbf{72.68} \\
        \bottomrule
    \end{tabular}
\end{table}

\paragraph{Models and Metrics.}

We evaluate OffQ on several widely adopted LLM families, including Llama~2 (7B/13B)~\cite{touvron2023llama}, Llama~3 (8B/70B)~\cite{meta2024llama3}, Llama~3.2 (1B/3B)~\cite{llama32_connect_2024}, and Qwen~2.5 (1.5B/3B/7B/ 14B/32B/72B)~\cite{qwen2025qwen25technicalreport}.
Model quality is assessed from two complementary perspectives: language modeling capability, measured by perplexity (PPL) on WikiText~\cite{merity2017pointer}, and common-sense reasoning, measured by zero-shot accuracy on ARC-e/ARC-c~\cite{clark2018thinksolvedquestionanswering}, BoolQ~\cite{clark-etal-2019-boolq}, HellaSwag~\cite{zellers2019hellaswag}, OpenBookQA~\cite{mihaylov-etal-2018-suit}, PIQA~\cite{Bisk2020}, SIQA~\cite{sap2019social}, and WinoGrande~\cite{sakaguchi2019winogrande}.

\paragraph{Baselines.}

We compare OffQ primarily against recent state-of-the-art rotation-based post-training quantization methods and their improvements, including QuaRot~\cite{ashkboos2024quarot}, SpinQuant~\cite{liu2025spinquant}, DFRot~\cite{xiang2025dfrot}, KurTail~\cite{akhondzadeh-etal-2025-kurtail}, OSTQuant~\cite{hu2025ostquant}, and ResQ~\cite{saxena2025resq}.
Specifically, SpinQuant, DFRot, and KurTail use learning based method for optimizing the rotation;
OSTQuant augments rotation with channel-wise scaling, yielding an affine transformation;
and ResQ retains high-precision for \(\nicefrac{1}{8}\) of the channels, resulting in an effective bit-width of 4.5.
For reference, we additionally include two representative PTQ baselines, the weight-only method GPTQ~\cite{frantar2023optq} and the mixed-precision method QUIK~\cite{QUIK}.

\paragraph{Implementation and Settings.}

We implement OffQ using the HuggingFace Transformers~\cite{wolf-etal-2020-transformers} and PyTorch~\cite{paszke2017automatic}, building upon the open-source codebases of ResQ.
For activations, OffQ adopts per-group asymmetric quantization with a group size of 128, whereas baselines follow their official implementations using per-token asymmetric quantization.
Weights are quantized with per-channel symmetric quantization across all methods.
Following the evaluation protocol of~\cite{saxena2025resq}, all methods except OSTQuant additionally apply GPTQ for weight quantization.
We focus on the W4A4KV4 setting, where weights, activations, and KV-cache are all quantized to 4 bits: a configuration that is both practically relevant and effective for reducing
memory footprint and bandwidth consumption.

\subsection{Quantization Results}

\cref{tab:result} summarizes results across ten LLMs of various sizes from both the Llama and Qwen families.
We make the following observations.
(1)~Weight-only methods, exemplified by GPTQ,
degrades severely under the W4A4KV4 setting: PPL increases from 6.1 to 166.3 on Llama 3-8B and from 6.8 to 13{,}594 on Qwen 2.5-7B.
(2)~The mixed-precision method QUIK substantially improves over GPTQ, yet still suffers from significant degradation.
(3)~Rotation-based methods achieve considerably better performance overall.
However, QuaRot fails on Qwen models smaller than 14B, suggesting that naive rotation alone is insufficient to suppress quantization error in the presence of severe outliers in small models.
(4)~Among rotation-based methods evaluated on the Llama family, learned-rotation approaches such as SpinQuant and KurTail improve over QuaRot but still exhibit a noticeable gap relative to the 16-bit baseline.
OSTQuant, which augments rotation with an additional scaling transformation, and ResQ, which enhances rotation with mixed-precision, close this gap further.
(5)~Our proposed OffQ achieves the best performance across nearly all models and benchmarks, with PPL closer to the full-precision baseline and zero-shot accuracy on par with or exceeding that of the best prior methods, without relying on mixed-precision computation or learned rotation.

\subsection{Ablation Study}

\begin{table}[t]
    \caption{
        Ablation studies on Llama 3-8B.
        The table shows the impact of each design choice on WikiText perplexity (PPL \(\downarrow\)) and the zero-shot accuracy (\(\uparrow\)) across 8 common-sense reasoning tasks.
    }\label{tab:ablation}
    \centering
    \setlength{\tabcolsep}{1.9pt}
    \begin{tabular}{llcccccccccccccc}
        \toprule
        \multirow{2.5}{*}{Ablation} & PPL  & \multicolumn{10}{c}{0-shot}                                                                          \\
        \cmidrule(r){2-2}\cmidrule(l){3-12}
                                    & Wiki & ARC-c                       & ARC-e & BoolQ & HellaS & OBQA  & PIQA  & SIQA  & WinoG & \textbf{Avg.} \\
        \midrule
        OffQ                        & 6.98 & 50.68                       & 77.44 & 80.43 & 76.96  & 43.80 & 78.89 & 45.96 & 69.77 & 65.49         \\
        \cdashline{1-15}\noalign{\vskip 3.5pt}
        Without Top-1 Selection     & 8.27 & 47.78                       & 73.99 & 77.95 & 74.24  & 41.20 & 77.58 & 43.09 & 69.22 & 63.13         \\
        Zigzag Grouping             & 7.03 & 49.40                       & 75.38 & 79.72 & 76.30  & 43.20 & 79.00 & 45.29 & 71.27 & 64.95         \\
        Partial Random Rotation     & 7.00 & 47.53                       & 74.33 & 79.33 & 76.40  & 42.00 & 79.60 & 45.80 & 70.24 & 64.40         \\
        \bottomrule
    \end{tabular}
\end{table}

We conduct ablation studies to analyze the contribution of each component in OffQ.
Starting from the full OffQ configuration, we modify each design choice individually.
The results are summarized in \cref{tab:ablation}, and we discuss the key findings from each ablation below.

\paragraph{Top-1 Selection in Activation Statistics.}

When computing activation statistics for outlier subspace identification, we select the token with the largest \(L^\infty\) norm within each sequence (top-1 selection) rather than using all tokens as in standard PCA~\cite{saxena2025resq}.
Comparing row~1 (with top-1 selection) and row~2 (without top-1 selection) of \cref{tab:ablation}, top-1 PCA substantially outperforms standard PCA with all tokens
(PPL: 6.98 \versus 8.27; 0-shot\(^8\): 65.49 \versus 63.13).
Together with the qualitative illustration in \cref{fig:pca}, this confirms that accurately identifying the outlier subspace, which requires focusing on the most extreme outlier tokens, is critical for effective outlier suppression in the offsetting technique.

\paragraph{Sorted \versus Zigzag Grouping.}

We compare the two strategies for grouping non-outlier channels.
Suppose we have 6 values \([1,2,3,4,50,60]\) with 2 outliers and 2 groups, sorted grouping yields \([60,1,2 ~|~ 50,3,4]\), while zigzag grouping yields \([60,1,3 ~|~ 50,2,4]\).
We vary the construction of permutation matrices \(P_1\) and \(P_4\) according to the two grouping strategies and find that zigzag grouping (row 3 of \cref{tab:ablation}) performs slightly worse than the sorted grouping in OffQ (PPL: 7.03 \versus 6.98; 0-shot\(^8\): 64.95 \versus 65.49), likely because sorted grouping clusters channels of similar variance, thereby reducing quantization error.

\paragraph{Hadamard \versus Partially Random Rotation.}

We constructed the partially random Hadamard matrix with all-one in the first row to achieve the offsetting effect, but it can alternatively be replaced by a general partially random rotation matrix whose first row is constant at \(\nicefrac{1}{\sqrt{D}}\).
On Llama 3-8B, substituting \(H_1\) with such a partially random rotation matrix (row 4 of \cref{tab:ablation}) yields comparable perplexity but lower zero-shot accuracy compared with OffQ (PPL: 7.00 \versus 6.98; 0-shot\(^8\): 64.40 \versus 65.49), suggesting the Hadamard structure provides a marginal benefit.

\paragraph{Number of Groups.}

\begin{figure}[t]
    \centering
    \includegraphics[width=0.59\textwidth]{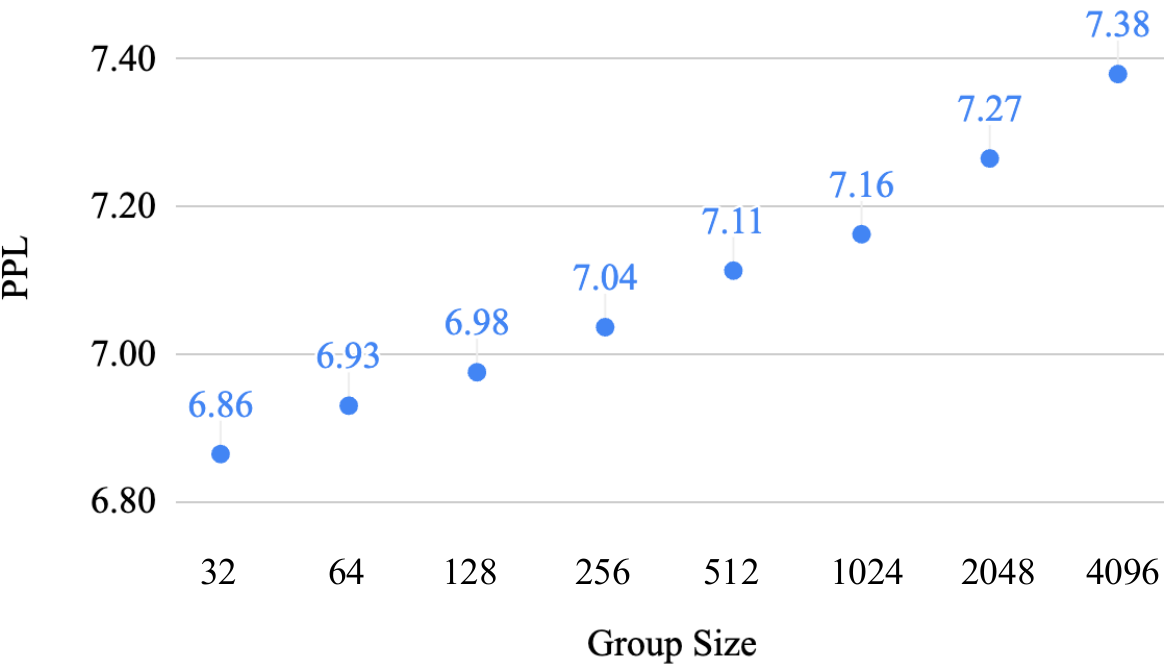}
    \caption{
        Effect of group size on quantization performance of OffQ on Llama 3-8B with a hidden dimension of 4096.
        Increasing the number of groups (\ie, reducing group size) leads to better quantization performance (lower perplexity) as more outlier channels can be absorbed by offsetting.
    }\label{fig:group_size}
\end{figure}

\cref{fig:group_size} shows the effect of group size on quantization performance.
We vary the group size for attention, feedforward, and down-projection inputs, while fixing the group size for the KV-cache to match the head dimension.
As expected, more groups allow more outlier channels to be absorbed, reducing quantization error and perplexity;
however, finer grouping also increases the number of quantization parameters (scales and zero-points) that must be stored.
We select a group size of 128 for activations (average bit-width of 4.25) as it strikes a good balance between quantization performance and storage overhead.

\section{Limitations}\label{sec:limitation}

OffQ achieves strong quantization performance under the W4A4KV4 setting, yet limitations remain.
First, while we focus on the effectiveness of OffQ in mitigating outlier-induced quantization error, we have not reported inference latency in real-world deployment scenarios.
Second, we leave the exploration of combining OffQ with complementary quantization techniques, such as non-uniform weight quantization~\cite{savkin2025nestquant,zandieh2026turboquant,choi-etal-2025-rotate}, advanced weight quantization methods~\cite{li2025gptaq,arai2025quantization,zhang2025leanquant,zhang2026qronos,li2026rethinking}, and rotation learning~\cite{xiang2025dfrot,akhondzadeh-etal-2025-kurtail,shao2026dartquant}, as future work.

\section{Conclusion}\label{sec:conclusion}

In this paper, we have presented OffQ, a post-training quantization method that addresses outlier-induced quantization error under the W4A4KV4 setting.
The core of OffQ is an offsetting technique that converts outlier activation channels into a uniform offset, which is absorbed into the zero-point of asymmetric quantization.
To effectively identify the outlier subspace and concentrate the outlier directions into the least number of channels, we have proposed a Top-1 PCA method, which efficiently captures the most significant variance in the activation distribution.
Extensive experiments on Llama and Qwen 2.5 series of LLMs have demonstrated that OffQ consistently outperforms state-of-the-art rotation-based PTQ methods across diverse model sizes and benchmarks, achieving superior perplexity and 0-shot accuracy while maintaining efficient memory usage.
These results highlight the effectiveness of OffQ in enabling high-performance LLM inference on resource-constrained hardware, paving the way for broader deployment of LLMs in real-world applications.

\begin{ack}
  This work was supported in part by the Swiss National Science Foundation via the grant 200020\_214878.
\end{ack}

\medskip

{
  \small
  \bibliographystyle{ieeenat_fullname}
  \bibliography{custom}

@inproceedings{wang2024sinder,
  title        = {Sinder: Repairing the singular defects of dinov2},
  author       = {Wang, Haoqi and Zhang, Tong and Salzmann, Mathieu},
  booktitle    = {European Conference on Computer Vision},
  pages        = {20--35},
  year         = {2024},
  organization = {Springer}
}

@inproceedings{wang2025demystifying,
  title     = {Demystifying Singular Defects in Large Language Models},
  author    = {Haoqi Wang and Tong Zhang and Mathieu Salzmann},
  booktitle = {Forty-second International Conference on Machine Learning},
  year      = {2025},
  url       = {https://openreview.net/forum?id=4yBnUokU2v}
}

@inproceedings{jacob2018quantization,
  title     = {Quantization and training of neural networks for efficient integer-arithmetic-only inference},
  author    = {Jacob, Benoit and Kligys, Skirmantas and Chen, Bo and Zhu, Menglong and Tang, Matthew and Howard, Andrew and Adam, Hartwig and Kalenichenko, Dmitry},
  booktitle = {Proceedings of the IEEE conference on computer vision and pattern recognition},
  pages     = {2704--2713},
  year      = {2018}
}

@inproceedings{frantar2023optq,
  title     = {{OPTQ}: Accurate Quantization for Generative Pre-trained Transformers},
  author    = {Elias Frantar and Saleh Ashkboos and Torsten Hoefler and Dan Alistarh},
  booktitle = {The Eleventh International Conference on Learning Representations },
  year      = {2023},
  url       = {https://openreview.net/forum?id=tcbBPnfwxS}
}

@misc{muller2025sinq,
  title         = {SINQ: Sinkhorn-Normalized Quantization for Calibration-Free Low-Precision LLM Weights},
  author        = {Lorenz K. Muller and Philippe Bich and Jiawei Zhuang and Ahmet Celik and Luca Benfenati and Lukas Cavigelli},
  year          = {2025},
  eprint        = {2509.22944},
  archiveprefix = {arXiv},
  primaryclass  = {cs.LG},
  url           = {http://arxiv.org/abs/2509.22944}
}

@inproceedings{liu2025spinquant,
  title     = {SpinQuant: {LLM} Quantization with Learned Rotations},
  author    = {Zechun Liu and Changsheng Zhao and Igor Fedorov and Bilge Soran and Dhruv Choudhary and Raghuraman Krishnamoorthi and Vikas Chandra and Yuandong Tian and Tijmen Blankevoort},
  booktitle = {The Thirteenth International Conference on Learning Representations},
  year      = {2025},
  url       = {https://openreview.net/forum?id=ogO6DGE6FZ}
}

@inproceedings{saxena2025resq,
  title     = {ResQ: Mixed-Precision Quantization of Large Language Models with Low-Rank Residuals},
  author    = {Utkarsh Saxena and Sayeh Sharify and Kaushik Roy and Xin Wang},
  booktitle = {Forty-second International Conference on Machine Learning},
  year      = {2025},
  url       = {https://openreview.net/forum?id=4qIP1sXcR1}
}

@inproceedings{chen-etal-2025-efficientqat,
  title     = {{E}fficient{QAT}: Efficient Quantization-Aware Training for Large Language Models},
  author    = {Chen, Mengzhao  and
               Shao, Wenqi  and
               Xu, Peng  and
               Wang, Jiahao  and
               Gao, Peng  and
               Zhang, Kaipeng  and
               Luo, Ping},
  editor    = {Che, Wanxiang  and
               Nabende, Joyce  and
               Shutova, Ekaterina  and
               Pilehvar, Mohammad Taher},
  booktitle = {Proceedings of the 63rd Annual Meeting of the Association for Computational Linguistics (Volume 1: Long Papers)},
  month     = jul,
  year      = {2025},
  address   = {Vienna, Austria},
  publisher = {Association for Computational Linguistics},
  url       = {https://aclanthology.org/2025.acl-long.498/},
  doi       = {10.18653/v1/2025.acl-long.498},
  pages     = {10081--10100},
  isbn      = {979-8-89176-251-0}
}

@inproceedings{liu-etal-2024-llm,
  title     = {{LLM}-{QAT}: Data-Free Quantization Aware Training for Large Language Models},
  author    = {Liu, Zechun  and
               Oguz, Barlas  and
               Zhao, Changsheng  and
               Chang, Ernie  and
               Stock, Pierre  and
               Mehdad, Yashar  and
               Shi, Yangyang  and
               Krishnamoorthi, Raghuraman  and
               Chandra, Vikas},
  editor    = {Ku, Lun-Wei  and
               Martins, Andre  and
               Srikumar, Vivek},
  booktitle = {Findings of the Association for Computational Linguistics: ACL 2024},
  month     = aug,
  year      = {2024},
  address   = {Bangkok, Thailand},
  publisher = {Association for Computational Linguistics},
  url       = {https://aclanthology.org/2024.findings-acl.26/},
  doi       = {10.18653/v1/2024.findings-acl.26},
  pages     = {467--484}
}

@inproceedings{ashkboos2025halo,
  title     = {{HALO}: Hadamard-Assisted Lower-Precision Optimization for {LLM}s},
  author    = {Saleh Ashkboos and Mahdi Nikdan and Soroush Tabesh and Roberto L. Castro and Torsten Hoefler and Dan Alistarh},
  booktitle = {The Thirty-ninth Annual Conference on Neural Information Processing Systems},
  year      = {2025},
  url       = {https://openreview.net/forum?id=OcMpSh79aE}
}

@inproceedings{xiao2023smoothquant,
  title     = {{S}mooth{Q}uant: Accurate and Efficient Post-Training Quantization for Large Language Models},
  author    = {Xiao, Guangxuan and Lin, Ji and Seznec, Mickael and Wu, Hao and Demouth, Julien and Han, Song},
  booktitle = {Proceedings of the 40th International Conference on Machine Learning},
  year      = {2023}
}

@inproceedings{wei-etal-2023-outlier,
  title     = {Outlier Suppression+: Accurate quantization of large language models by equivalent and effective shifting and scaling},
  author    = {Wei, Xiuying  and
               Zhang, Yunchen  and
               Li, Yuhang  and
               Zhang, Xiangguo  and
               Gong, Ruihao  and
               Guo, Jinyang  and
               Liu, Xianglong},
  editor    = {Bouamor, Houda  and
               Pino, Juan  and
               Bali, Kalika},
  booktitle = {Proceedings of the 2023 Conference on Empirical Methods in Natural Language Processing},
  month     = dec,
  year      = {2023},
  address   = {Singapore},
  publisher = {Association for Computational Linguistics},
  url       = {https://aclanthology.org/2023.emnlp-main.102/},
  doi       = {10.18653/v1/2023.emnlp-main.102},
  pages     = {1648--1665}
}

@inproceedings{shao2024omniquant,
  title     = {OmniQuant: Omnidirectionally Calibrated Quantization for Large Language Models},
  author    = {Wenqi Shao and Mengzhao Chen and Zhaoyang Zhang and Peng Xu and Lirui Zhao and Zhiqian Li and Kaipeng Zhang and Peng Gao and Yu Qiao and Ping Luo},
  booktitle = {The Twelfth International Conference on Learning Representations},
  year      = {2024},
  url       = {https://openreview.net/forum?id=8Wuvhh0LYW}
}

@misc{chee2024quip2bitquantizationlarge,
  title         = {QuIP: 2-Bit Quantization of Large Language Models With Guarantees},
  author        = {Jerry Chee and Yaohui Cai and Volodymyr Kuleshov and Christopher De Sa},
  year          = {2024},
  eprint        = {2307.13304},
  archiveprefix = {arXiv},
  primaryclass  = {cs.LG},
  url           = {https://arxiv.org/abs/2307.13304}
}

@inproceedings{tseng2024quip,
  title     = {Qu{IP}\${\textbackslash}\#\$: Even Better {LLM} Quantization with Hadamard Incoherence and Lattice Codebooks},
  author    = {Albert Tseng and Jerry Chee and Qingyao Sun and Volodymyr Kuleshov and Christopher De Sa},
  booktitle = {Forty-first International Conference on Machine Learning},
  year      = {2024},
  url       = {https://openreview.net/forum?id=9BrydUVcoe}
}

@article{ashkboos2024quarot,
  title   = {Quarot: Outlier-free 4-bit inference in rotated llms},
  author  = {Ashkboos, Saleh and Mohtashami, Amirkeivan and Croci, Maximilian L and Li, Bo and Cameron, Pashmina and Jaggi, Martin and Alistarh, Dan and Hoefler, Torsten and Hensman, James},
  journal = {Advances in Neural Information Processing Systems},
  volume  = {37},
  pages   = {100213--100240},
  year    = {2024}
}

@inproceedings{hu2025ostquant,
  title     = {{OSTQ}uant: Refining Large Language Model Quantization with Orthogonal and Scaling Transformations for Better Distribution Fitting},
  author    = {Xing Hu and Yuan Cheng and Dawei Yang and Zhixuan Chen and Zukang Xu and JiangyongYu and XUCHEN and Zhihang Yuan and Zhe jiang and Sifan Zhou},
  booktitle = {The Thirteenth International Conference on Learning Representations},
  year      = {2025},
  url       = {https://openreview.net/forum?id=rAcgDBdKnP}
}

@inproceedings{shao2026dartquant,
  title     = {DartQuant: Efficient Rotational Distribution Calibration for {LLM} Quantization},
  author    = {Yuantian Shao and Yuanteng Chen and Peisong Wang and Jianlin Yu and Jing Lin and Yiwu Yao and Zhihui Wei and Jian Cheng},
  booktitle = {The Thirty-ninth Annual Conference on Neural Information Processing Systems},
  year      = {2026},
  url       = {https://openreview.net/forum?id=LfcfwlLCHM}
}

@inproceedings{he2024zipcache,
  title     = {ZipCache: Accurate and Efficient {KV} Cache Quantization with Salient Token Identification},
  author    = {Yefei He and Luoming Zhang and Weijia Wu and Jing Liu and Hong Zhou and Bohan Zhuang},
  booktitle = {The Thirty-eighth Annual Conference on Neural Information Processing Systems},
  year      = {2024},
  url       = {https://openreview.net/forum?id=5t4ZAkPiJs}
}

@inproceedings{hooper2024kvquant,
  title     = {{KVQ}uant: Towards 10 Million Context Length {LLM} Inference with {KV} Cache Quantization},
  author    = {Coleman Richard Charles Hooper and Sehoon Kim and Hiva Mohammadzadeh and Michael W. Mahoney and Sophia Shao and Kurt Keutzer and Amir Gholami},
  booktitle = {The Thirty-eighth Annual Conference on Neural Information Processing Systems},
  year      = {2024},
  url       = {https://openreview.net/forum?id=0LXotew9Du}
}

@inproceedings{liu2024kivi,
  title        = {KIVI: A Tuning-Free Asymmetric 2bit Quantization for KV Cache},
  author       = {Liu, Zirui and Yuan, Jiayi and Jin, Hongye and Zhong, Shaochen and Xu, Zhaozhuo and Braverman, Vladimir and Chen, Beidi and Hu, Xia},
  booktitle    = {International Conference on Machine Learning},
  pages        = {32332--32344},
  year         = {2024},
  organization = {PMLR}
}

@inproceedings{MLSYS2024_5edb57c0,
  author    = {Zhao, Yilong and Lin, Chien-Yu and Zhu, Kan and Ye, Zihao and Chen, Lequn and Zheng, Size and Ceze, Luis and Krishnamurthy, Arvind and Chen, Tianqi and Kasikci, Baris},
  booktitle = {Proceedings of Machine Learning and Systems},
  editor    = {P. Gibbons and G. Pekhimenko and C. De Sa},
  pages     = {196--209},
  title     = {Atom: Low-Bit Quantization for Efficient and Accurate LLM Serving},
  url       = {https://proceedings.mlsys.org/paper_files/paper/2024/file/5edb57c05c81d04beb716ef1d542fe9e-Paper-Conference.pdf},
  volume    = {6},
  year      = {2024}
}

@inproceedings{arai2025quantization,
  title     = {Quantization Error Propagation: Revisiting Layer-Wise Post-Training Quantization},
  author    = {Yamato Arai and Yuma Ichikawa},
  booktitle = {The Thirty-ninth Annual Conference on Neural Information Processing Systems},
  year      = {2025},
  url       = {https://openreview.net/forum?id=a3l3K9khbL}
}

@inproceedings{lin2023awq,
  title     = {AWQ: Activation-aware Weight Quantization for LLM Compression and Acceleration},
  author    = {Lin, Ji and Tang, Jiaming and Tang, Haotian and Yang, Shang and Chen, Wei-Ming and Wang, Wei-Chen and Xiao, Guangxuan and Dang, Xingyu and Gan, Chuang and Han, Song},
  booktitle = {MLSys},
  year      = {2024}
}

@inproceedings{onebit2024xu,
  title     = {OneBit: Towards Extremely Low-bit Large Language Models},
  author    = {Xu, Yuzhuang and Han, Xu and Yang, Zonghan and Wang, Shuo and Zhu, Qingfu and Liu, Zhiyuan and Liu, Weidong and Che, Wanxiang},
  booktitle = {Advances in Neural Information Processing Systems},
  year      = {2024},
  volume    = {37},
  pages     = {66357--66382}
}

@misc{2310.11453,
  author = {Hongyu Wang and Shuming Ma and Li Dong and Shaohan Huang and Huaijie Wang and Lingxiao Ma and Fan Yang and Ruiping Wang and Yi Wu and Furu Wei},
  title  = {BitNet: Scaling 1-bit Transformers for Large Language Models},
  year   = {2023},
  eprint = {arXiv:2310.11453}
}

@inproceedings{son2025nsnquant,
  title     = {{NSNQ}uant: A Double Normalization Approach for Calibration-Free Low-Bit Vector Quantization of {KV} Cache},
  author    = {Donghyun Son and Euntae Choi and Sungjoo Yoo},
  booktitle = {The Thirty-ninth Annual Conference on Neural Information Processing Systems},
  year      = {2025},
  url       = {https://openreview.net/forum?id=boNYskaXnO}
}

@article{vanbaalen-gptvq,
  title   = {GPTVQ: The Blessing of Dimensionality in LLM Quantization},
  author  = {Mart van Baalen and Andrey Kuzmin and Markus Nagel and Peter Couperus and Cedric Bastoul and Eric Mahurin and Tijmen Blankevoort and Paul Whatmough},
  year    = {2024},
  journal = {arXiv preprint arXiv:2402.15319}
}

@inproceedings{vptq,
  title     = {VPTQ: Extreme Low-bit Vector Post-Training Quantization for Large Language Models},
  author    = {Yifei Liu and
               Jicheng Wen and
               Yang Wang and
               Shengyu Ye and
               Li Lyna Zhang and
               Ting Cao and
               Cheng Li and
               Mao Yang},
  booktitle = {The 2024 Conference on Empirical Methods in Natural Language Processing},
  year      = {2024}
}

@article{xu2025crvq,
  title   = {{CRVQ}: Channel-Relaxed Vector Quantization for Extreme Compression of {LLMs}},
  author  = {Xu, Yuzhuang and Ji, Shiyu and Zhu, Qingfu and Che, Wanxiang},
  journal = {Transactions of the Association for Computational Linguistics (TACL)},
  volume  = {13},
  pages   = {1488-1506},
  year    = {2025},
  url     = {https://doi.org/10.1162/TACL.a.45}
}

@inproceedings{ma2024affinequant,
  title     = {AffineQuant: Affine Transformation Quantization for Large Language Models},
  author    = {Yuexiao Ma and Huixia Li and Xiawu Zheng and Feng Ling and Xuefeng Xiao and Rui Wang and Shilei Wen and Fei Chao and Rongrong Ji},
  booktitle = {The Twelfth International Conference on Learning Representations},
  year      = {2024},
  url       = {https://openreview.net/forum?id=of2rhALq8l}
}

@inproceedings{adepuFQIcml24,
  author          = {Harshavardhan Adepu and Zhanpeng Zeng and Li Zhang and Vikas Singh},
  title           = {FrameQuant: Flexible Low-Bit Quantization for Transformers},
  optcrossref     = {},
  optkey          = {},
  booktitle       = {Proceedings of International Conference on Machine Learning (ICML)},
  optpages        = {},
  year            = {2024},
  venue           = {ICML},
  opteditor       = {},
  optvolume       = {},
  optnumber       = {},
  optseries       = {},
  optaddress      = {},
  month           = {July},
  optorganization = {},
  optpublisher    = {},
  optnote         = {},
  optannote       = {}
}

@misc{kim2024squeezellm,
  title  = {Squeeze{LLM}: Dense and Sparse Quantization},
  author = {Sehoon Kim and Coleman Richard Charles Hooper and Amir Gholami and Zhen Dong and Xiuyu Li and Sheng Shen and Michael W. Mahoney and Kurt Keutzer},
  year   = {2024},
  url    = {https://openreview.net/forum?id=pZhdz4oyzo}
}

@inproceedings{du-etal-2024-bitdistiller,
  title     = {{B}it{D}istiller: Unleashing the Potential of Sub-4-Bit {LLM}s via Self-Distillation},
  author    = {Du, DaYou  and
               Zhang, Yijia  and
               Cao, Shijie  and
               Guo, Jiaqi  and
               Cao, Ting  and
               Chu, Xiaowen  and
               Xu, Ningyi},
  editor    = {Ku, Lun-Wei  and
               Martins, Andre  and
               Srikumar, Vivek},
  booktitle = {Proceedings of the 62nd Annual Meeting of the Association for Computational Linguistics (Volume 1: Long Papers)},
  month     = aug,
  year      = {2024},
  address   = {Bangkok, Thailand},
  publisher = {Association for Computational Linguistics},
  url       = {https://aclanthology.org/2024.acl-long.7/},
  doi       = {10.18653/v1/2024.acl-long.7},
  pages     = {102--116}
}

@article{wei2022outlier,
  title   = {Outlier suppression: Pushing the limit of low-bit transformer language models},
  author  = {Wei, Xiuying and Zhang, Yunchen and Zhang, Xiangguo and Gong, Ruihao and Zhang, Shanghang and Zhang, Qi and Yu, Fengwei and Liu, Xianglong},
  journal = {Advances in Neural Information Processing Systems},
  volume  = {35},
  pages   = {17402--17414},
  year    = {2022}
}

@article{dettmers2022llmint8,
  title   = {LLM.int8(): 8-bit matrix multiplication for transformers at scale},
  author  = {Dettmers, Tim and Lewis, Mike and Belkada, Younes and Zettlemoyer, Luke},
  journal = {Advances in neural information processing systems},
  volume  = {35},
  pages   = {30318--30332},
  year    = {2022}
}

@article{huang2024billm,
  title   = {BiLLM: Pushing the Limit of Post-Training Quantization for LLMs},
  author  = {Huang, Wei and Liu, Yangdong and Qin, Haotong and Li, Ying and Zhang, Shiming and Liu, Xianglong and Magno, Michele and Qi, Xiaojuan},
  journal = {arXiv preprint arXiv:2402.04291},
  year    = {2024}
}

@inproceedings{li2025gptaq,
  title     = {{GPTAQ}: Efficient Finetuning-Free Quantization for Asymmetric Calibration},
  author    = {Yuhang Li and Ruokai Yin and Donghyun Lee and Shiting Xiao and Priyadarshini Panda},
  booktitle = {Forty-second International Conference on Machine Learning},
  year      = {2025},
  url       = {https://openreview.net/forum?id=QdELyl0FST}
}

@misc{yuan2023rptq,
  title         = {RPTQ: Reorder-based Post-training Quantization for Large Language Models},
  author        = {Zhihang Yuan and Lin Niu and Jiawei Liu and Wenyu Liu and Xinggang Wang and Yuzhang Shang and Guangyu Sun and Qiang Wu and Jiaxiang Wu and Bingzhe Wu},
  year          = {2023},
  eprint        = {2304.01089},
  archiveprefix = {arXiv},
  primaryclass  = {cs.CL}
}

@inproceedings{liu2024qllm,
  title     = {{QLLM}: Accurate and Efficient Low-Bitwidth Quantization for Large Language Models},
  author    = {Liu, Jing and Gong, Ruihao and Wei, Xiuying and Dong, Zhiwei and Cai, Jianfei and Zhuang, Bohan},
  booktitle = {International Conference on Learning Representations (ICLR)},
  year      = {2024}
}

@article{lin2025qserve,
  title   = {Qserve: W4a8kv4 quantization and system co-design for efficient llm serving},
  author  = {Lin, Yujun and Tang, Haotian and Yang, Shang and Zhang, Zhekai and Xiao, Guangxuan and Gan, Chuang and Han, Song},
  journal = {Proceedings of Machine Learning and Systems},
  volume  = {7},
  year    = {2025}
}

@article{lin2024duquant,
  title   = {Duquant: Distributing outliers via dual transformation makes stronger quantized llms},
  author  = {Lin, Haokun and Xu, Haobo and Wu, Yichen and Cui, Jingzhi and Zhang, Yingtao and Mou, Linzhan and Song, Linqi and Sun, Zhenan and Wei, Ying},
  journal = {Advances in Neural Information Processing Systems},
  volume  = {37},
  pages   = {87766--87800},
  year    = {2024}
}

@inproceedings{dettmers2024spqr,
  title     = {Sp{QR}: A Sparse-Quantized Representation for Near-Lossless {LLM} Weight Compression},
  author    = {Tim Dettmers and Ruslan A. Svirschevski and Vage Egiazarian and Denis Kuznedelev and Elias Frantar and Saleh Ashkboos and Alexander Borzunov and Torsten Hoefler and Dan Alistarh},
  booktitle = {The Twelfth International Conference on Learning Representations},
  year      = {2024},
  url       = {https://openreview.net/forum?id=Q1u25ahSuy}
}

@inproceedings{pmlr-v235-kim24f,
  title     = {{S}queeze{LLM}: Dense-and-Sparse Quantization},
  author    = {Kim, Sehoon and Hooper, Coleman Richard Charles and Gholami, Amir and Dong, Zhen and Li, Xiuyu and Shen, Sheng and Mahoney, Michael W. and Keutzer, Kurt},
  booktitle = {Proceedings of the 41st International Conference on Machine Learning},
  pages     = {23901--23923},
  year      = {2024},
  editor    = {Salakhutdinov, Ruslan and Kolter, Zico and Heller, Katherine and Weller, Adrian and Oliver, Nuria and Scarlett, Jonathan and Berkenkamp, Felix},
  volume    = {235},
  series    = {Proceedings of Machine Learning Research},
  month     = {21--27 Jul},
  publisher = {PMLR},
  url       = {https://proceedings.mlr.press/v235/kim24f.html}
}

@inproceedings{zeng2025abq,
  title     = {Abq-llm: Arbitrary-bit quantized inference acceleration for large language models},
  author    = {Zeng, Chao and Liu, Songwei and Xie, Yusheng and Liu, Hong and Wang, Xiaojian and Wei, Miao and Yang, Shu and Chen, Fangmin and Mei, Xing},
  booktitle = {Proceedings of the AAAI Conference on Artificial Intelligence},
  volume    = {39},
  number    = {21},
  pages     = {22299--22307},
  year      = {2025}
}

@inproceedings{QUIK,
  title     = {{QUIK}: Towards End-to-end 4-Bit Inference on Generative Large Language Models},
  author    = {Ashkboos, Saleh  and
               Markov, Ilia  and
               Frantar, Elias  and
               Zhong, Tingxuan  and
               Wang, Xincheng  and
               Ren, Jie  and
               Hoefler, Torsten  and
               Alistarh, Dan},
  editor    = {Al-Onaizan, Yaser  and
               Bansal, Mohit  and
               Chen, Yun-Nung},
  booktitle = {Proceedings of the 2024 Conference on Empirical Methods in Natural Language Processing},
  month     = nov,
  year      = {2024},
  address   = {Miami, Florida, USA},
  publisher = {Association for Computational Linguistics},
  url       = {https://aclanthology.org/2024.emnlp-main.197/},
  doi       = {10.18653/v1/2024.emnlp-main.197},
  pages     = {3355--3371}
}

@inproceedings{pmlr-v267-huang25aa,
  title     = {{S}li{M}-{LLM}: Salience-Driven Mixed-Precision Quantization for Large Language Models},
  author    = {Huang, Wei and Qin, Haotong and Liu, Yangdong and Li, Yawei and Liu, Qinshuo and Liu, Xianglong and Benini, Luca and Magno, Michele and Zhang, Shiming and Qi, Xiaojuan},
  booktitle = {Proceedings of the 42nd International Conference on Machine Learning},
  pages     = {25672--25692},
  year      = {2025},
  editor    = {Singh, Aarti and Fazel, Maryam and Hsu, Daniel and Lacoste-Julien, Simon and Berkenkamp, Felix and Maharaj, Tegan and Wagstaff, Kiri and Zhu, Jerry},
  volume    = {267},
  series    = {Proceedings of Machine Learning Research},
  month     = {13--19 Jul},
  publisher = {PMLR}
}

@inproceedings{kim-etal-2022-understanding,
  title     = {Understanding and Improving Knowledge Distillation for Quantization Aware Training of Large Transformer Encoders},
  author    = {Kim, Minsoo  and
               Lee, Sihwa  and
               Hong, Suk-Jin  and
               Chang, Du-Seong  and
               Choi, Jungwook},
  editor    = {Goldberg, Yoav  and
               Kozareva, Zornitsa  and
               Zhang, Yue},
  booktitle = {Proceedings of the 2022 Conference on Empirical Methods in Natural Language Processing},
  month     = dec,
  year      = {2022},
  address   = {Abu Dhabi, United Arab Emirates},
  publisher = {Association for Computational Linguistics},
  url       = {https://aclanthology.org/2022.emnlp-main.450/},
  doi       = {10.18653/v1/2022.emnlp-main.450},
  pages     = {6713--6725}
}

@inproceedings{oneill-dutta-2023-self,
  title     = {Self-Distilled Quantization: Achieving High Compression Rates in Transformer-Based Language Models},
  author    = {O{'}Neill, James  and
               Dutta, Sourav},
  editor    = {Rogers, Anna  and
               Boyd-Graber, Jordan  and
               Okazaki, Naoaki},
  booktitle = {Proceedings of the 61st Annual Meeting of the Association for Computational Linguistics (Volume 2: Short Papers)},
  month     = jul,
  year      = {2023},
  address   = {Toronto, Canada},
  publisher = {Association for Computational Linguistics},
  url       = {https://aclanthology.org/2023.acl-short.114/},
  doi       = {10.18653/v1/2023.acl-short.114},
  pages     = {1329--1339}
}

@article{touvron2023llama,
  title={Llama 2: Open foundation and fine-tuned chat models},
  author={Touvron, Hugo and Martin, Louis and Stone, Kevin and Albert, Peter and Almahairi, Amjad and Babaei, Yasmine and Bashlykov, Nikolay and Batra, Soumya and Bhargava, Prajjwal and Bhosale, Shruti and others},
  journal={arXiv preprint arXiv:2307.09288},
  year={2023}
}

@article{meta2024llama3,
  title={The llama 3 herd of models},
  author={Grattafiori, Aaron and Dubey, Abhimanyu and Jauhri, Abhinav and Pandey, Abhinav and Kadian, Abhishek and Al-Dahle, Ahmad and Letman, Aiesha and Mathur, Akhil and Schelten, Alan and Vaughan, Alex and others},
  journal={arXiv preprint arXiv:2407.21783},
  year={2024}
}

@misc{llama32_connect_2024,
  title        = {Llama 3.2: Revolutionizing edge AI and vision with open, customizable models},
  author       = {{Meta AI}},
  year         = {2024},
  month        = sep,
  day          = {25},
  howpublished = {\url{https://ai.meta.com/blog/llama-3-2-connect-2024-vision-edge-mobile-devices/}},
  note         = {Accessed: 2026-05-05}
}

@misc{qwen2025qwen25technicalreport,
  title         = {Qwen2.5 Technical Report},
  author        = {Qwen and : and An Yang and Baosong Yang and Beichen Zhang and Binyuan Hui and Bo Zheng and Bowen Yu and Chengyuan Li and Dayiheng Liu and Fei Huang and Haoran Wei and Huan Lin and Jian Yang and Jianhong Tu and Jianwei Zhang and Jianxin Yang and Jiaxi Yang and Jingren Zhou and Junyang Lin and Kai Dang and Keming Lu and Keqin Bao and Kexin Yang and Le Yu and Mei Li and Mingfeng Xue and Pei Zhang and Qin Zhu and Rui Men and Runji Lin and Tianhao Li and Tianyi Tang and Tingyu Xia and Xingzhang Ren and Xuancheng Ren and Yang Fan and Yang Su and Yichang Zhang and Yu Wan and Yuqiong Liu and Zeyu Cui and Zhenru Zhang and Zihan Qiu},
  year          = {2025},
  eprint        = {2412.15115},
  archiveprefix = {arXiv},
  primaryclass  = {cs.CL},
  url           = {https://arxiv.org/abs/2412.15115}
}

@inproceedings{merity2017pointer,
  title     = {Pointer Sentinel Mixture Models},
  author    = {Stephen Merity and Caiming Xiong and James Bradbury and Richard Socher},
  booktitle = {International Conference on Learning Representations},
  year      = {2017},
  url       = {https://openreview.net/forum?id=Byj72udxe}
}

@misc{clark2018thinksolvedquestionanswering,
  title         = {Think you have Solved Question Answering? Try ARC, the AI2 Reasoning Challenge},
  author        = {Peter Clark and Isaac Cowhey and Oren Etzioni and Tushar Khot and Ashish Sabharwal and Carissa Schoenick and Oyvind Tafjord},
  year          = {2018},
  eprint        = {1803.05457},
  archiveprefix = {arXiv},
  primaryclass  = {cs.AI},
  url           = {https://arxiv.org/abs/1803.05457}
}

@inproceedings{clark-etal-2019-boolq,
  title     = {{B}ool{Q}: Exploring the Surprising Difficulty of Natural Yes/No Questions},
  author    = {Clark, Christopher  and
               Lee, Kenton  and
               Chang, Ming-Wei  and
               Kwiatkowski, Tom  and
               Collins, Michael  and
               Toutanova, Kristina},
  editor    = {Burstein, Jill  and
               Doran, Christy  and
               Solorio, Thamar},
  booktitle = {Proceedings of the 2019 Conference of the North {A}merican Chapter of the Association for Computational Linguistics: Human Language Technologies, Volume 1 (Long and Short Papers)},
  month     = jun,
  year      = {2019},
  address   = {Minneapolis, Minnesota},
  publisher = {Association for Computational Linguistics},
  url       = {https://aclanthology.org/N19-1300/},
  doi       = {10.18653/v1/N19-1300},
  pages     = {2924--2936}
}

@inproceedings{zellers2019hellaswag,
  title     = {HellaSwag: Can a Machine Really Finish Your Sentence?},
  author    = {Zellers, Rowan and Holtzman, Ari and Bisk, Yonatan and Farhadi, Ali and Choi, Yejin},
  booktitle = {Proceedings of the 57th Annual Meeting of the Association for Computational Linguistics},
  year      = {2019}
}

@inproceedings{mihaylov-etal-2018-suit,
  title     = {Can a Suit of Armor Conduct Electricity? A New Dataset for Open Book Question Answering},
  author    = {Mihaylov, Todor  and
               Clark, Peter  and
               Khot, Tushar  and
               Sabharwal, Ashish},
  editor    = {Riloff, Ellen  and
               Chiang, David  and
               Hockenmaier, Julia  and
               Tsujii, Jun{'}ichi},
  booktitle = {Proceedings of the 2018 Conference on Empirical Methods in Natural Language Processing},
  month     = oct # {-} # nov,
  year      = {2018},
  address   = {Brussels, Belgium},
  publisher = {Association for Computational Linguistics},
  url       = {https://aclanthology.org/D18-1260/},
  doi       = {10.18653/v1/D18-1260},
  pages     = {2381--2391}
}

@inproceedings{Bisk2020,
  author    = {Yonatan Bisk and Rowan Zellers and
               Ronan Le Bras and Jianfeng Gao
               and Yejin Choi},
  title     = {PIQA: Reasoning about Physical Commonsense in
               Natural Language},
  booktitle = {Thirty-Fourth AAAI Conference on
               Artificial Intelligence},
  year      = {2020}
}

@inproceedings{sap2019social,
  title     = {Social IQa: Commonsense Reasoning about Social Interactions},
  author    = {Sap, Maarten and Rashkin, Hannah and Chen, Derek and Le Bras, Ronan and Choi, Yejin},
  booktitle = {Proceedings of the 2019 Conference on Empirical Methods in Natural Language Processing and the 9th International Joint Conference on Natural Language Processing (EMNLP-IJCNLP)},
  pages     = {4463--4473},
  year      = {2019}
}

@article{sakaguchi2019winogrande,
  title   = {WinoGrande: An Adversarial Winograd Schema Challenge at Scale},
  author  = {Sakaguchi, Keisuke and Bras, Ronan Le and Bhagavatula, Chandra and Choi, Yejin},
  journal = {arXiv preprint arXiv:1907.10641},
  year    = {2019}
}

@inproceedings{sun2025flatquant,
  title     = {FlatQuant: Flatness Matters for {LLM} Quantization},
  author    = {Yuxuan Sun and Ruikang Liu and Haoli Bai and Han Bao and Kang Zhao and Yuening Li and JiaxinHu and Xianzhi Yu and Lu Hou and Chun Yuan and Xin Jiang and Wulong Liu and Jun Yao},
  booktitle = {Forty-second International Conference on Machine Learning},
  year      = {2025},
  url       = {https://openreview.net/forum?id=uTz2Utym5n}
}

@inproceedings{zandieh2026turboquant,
  title     = {TurboQuant: Online Vector Quantization with Near-optimal Distortion Rate},
  author    = {Amir Zandieh and Majid Daliri and Majid Hadian and Vahab Mirrokni},
  booktitle = {The Fourteenth International Conference on Learning Representations},
  year      = {2026},
  url       = {https://openreview.net/forum?id=tO3ASKZlok}
}

@inproceedings{choi-etal-2025-rotate,
  title     = {Rotate, Clip, and Partition: Towards {W}2{A}4{KV}4 Quantization by Integrating Rotation and Learnable Non-uniform Quantizer},
  author    = {Choi, Euntae  and
               Song, Sumin  and
               Lim, Woosang  and
               Yoo, Sungjoo},
  editor    = {Christodoulopoulos, Christos  and
               Chakraborty, Tanmoy  and
               Rose, Carolyn  and
               Peng, Violet},
  booktitle = {Findings of the Association for Computational Linguistics: EMNLP 2025},
  month     = nov,
  year      = {2025},
  address   = {Suzhou, China},
  publisher = {Association for Computational Linguistics},
  url       = {https://aclanthology.org/2025.findings-emnlp.400/},
  doi       = {10.18653/v1/2025.findings-emnlp.400},
  pages     = {7568--7590},
  isbn      = {979-8-89176-335-7}
}

@inproceedings{bondarenko2021understanding,
  title     = {Understanding and overcoming the challenges of efficient transformer quantization},
  author    = {Bondarenko, Yelysei and Nagel, Markus and Blankevoort, Tijmen},
  booktitle = {Proceedings of the 2021 Conference on Empirical Methods in Natural Language Processing},
  pages     = {7947--7969},
  year      = {2021}
}

@inproceedings{sun2024massive,
  title     = {Massive Activations in Large Language Models},
  author    = {Mingjie Sun and Xinlei Chen and J Zico Kolter and Zhuang Liu},
  booktitle = {First Conference on Language Modeling},
  year      = {2024},
  url       = {https://openreview.net/forum?id=F7aAhfitX6}
}

@inproceedings{akhondzadeh-etal-2025-kurtail,
  title     = {{K}ur{T}ail : Kurtosis-based {LLM} Quantization},
  author    = {Akhondzadeh, Mohammad Sadegh  and
               Bojchevski, Aleksandar  and
               Eleftheriou, Evangelos  and
               Dazzi, Martino},
  editor    = {Christodoulopoulos, Christos  and
               Chakraborty, Tanmoy  and
               Rose, Carolyn  and
               Peng, Violet},
  booktitle = {Findings of the Association for Computational Linguistics: EMNLP 2025},
  month     = nov,
  year      = {2025},
  address   = {Suzhou, China},
  publisher = {Association for Computational Linguistics},
  url       = {https://aclanthology.org/2025.findings-emnlp.943/},
  doi       = {10.18653/v1/2025.findings-emnlp.943},
  pages     = {17404--17419},
  isbn      = {979-8-89176-335-7}
}

@article{tseng2024qtip,
  title   = {Qtip: Quantization with trellises and incoherence processing},
  author  = {Tseng, Albert and Sun, Qingyao and Hou, David and De, Christopher},
  journal = {Advances in Neural Information Processing Systems},
  volume  = {37},
  pages   = {59597--59620},
  year    = {2024}
}

@inproceedings{su2025accurate,
  title     = {Accurate kv cache quantization with outlier tokens tracing},
  author    = {Su, Yi and Zhou, Yuechi and Qiu, Quantong and Li, Juntao and Xia, Qingrong and Li, Ping and Duan, Xinyu and Wang, Zhefeng and Zhang, Min},
  booktitle = {Proceedings of the 63rd Annual Meeting of the Association for Computational Linguistics (Volume 1: Long Papers)},
  pages     = {12895--12915},
  year      = {2025}
}

@inproceedings{zhang2024afpq,
  title     = {Afpq: Asymmetric floating point quantization for llms},
  author    = {Zhang, Yijia and Zhang, Sicheng and Cao, Shijie and Du, Dayou and Wei, Jianyu and Cao, Ting and Xu, Ningyi},
  booktitle = {Findings of the Association for Computational Linguistics: ACL 2024},
  pages     = {28--36},
  year      = {2024}
}

@inproceedings{hu2025mlwq,
  title     = {MLWQ: Efficient Small Language Model Deployment via Multi-Level Weight Quantization},
  author    = {Hu, Chun and He, Junhui and Wu, Shangyu and He, Yuxin and Xue, Chun Jason and Li, Qingan},
  booktitle = {Proceedings of the 2025 Conference on Empirical Methods in Natural Language Processing},
  pages     = {8078--8088},
  year      = {2025}
}

@inproceedings{guo2024lqlora,
  title     = {{LQ}-Lo{RA}: Low-rank plus Quantized Matrix Decomposition for Efficient Language Model Finetuning},
  author    = {Han Guo and Philip Greengard and Eric Xing and Yoon Kim},
  booktitle = {The Twelfth International Conference on Learning Representations},
  year      = {2024},
  url       = {https://openreview.net/forum?id=xw29VvOMmU}
}

@inproceedings{zhang2025leanquant,
  title     = {LeanQuant: Accurate and Scalable Large Language Model Quantization with Loss-error-aware Grid},
  author    = {Tianyi Zhang and Anshumali Shrivastava},
  booktitle = {The Thirteenth International Conference on Learning Representations},
  year      = {2025},
  url       = {https://openreview.net/forum?id=ISqx8giekS}
}

@inproceedings{lee2023flexround,
  title        = {Flexround: Learnable rounding based on element-wise division for post-training quantization},
  author       = {Lee, Jung Hyun and Kim, Jeonghoon and Kwon, Se Jung and Lee, Dongsoo},
  booktitle    = {International Conference on Machine Learning},
  pages        = {18913--18939},
  year         = {2023},
  organization = {PMLR}
}

@article{zhang2024magr,
  title   = {Magr: Weight magnitude reduction for enhancing post-training quantization},
  author  = {Zhang, Aozhong and Wang, Naigang and Deng, Yanxia and Li, Xin and Yang, Zi and Yin, Penghang},
  journal = {Advances in neural information processing systems},
  volume  = {37},
  pages   = {85109--85130},
  year    = {2024}
}

@inproceedings{arai2026quantization,
  title     = {Quantization Error Propagation: Revisiting Layer-Wise Post-Training Quantization},
  author    = {Yamato Arai and Yuma Ichikawa},
  booktitle = {The Thirty-ninth Annual Conference on Neural Information Processing Systems},
  year      = {2026},
  url       = {https://openreview.net/forum?id=a3l3K9khbL}
}

@inproceedings{cai2025pyramidkv,
  title     = {Pyramid{KV}: Dynamic {KV} Cache Compression based on Pyramidal Information Funneling},
  author    = {Zefan Cai and Yichi Zhang and Bofei Gao and Yuliang Liu and Yucheng Li and Tianyu Liu and Keming Lu and Wayne Xiong and Yue Dong and Junjie Hu and Wen Xiao},
  booktitle = {Second Conference on Language Modeling},
  year      = {2025},
  url       = {https://openreview.net/forum?id=ayi7qezU87}
}

@inproceedings{yang-etal-2025-xquant,
  title     = {{XQ}uant: Achieving Ultra-Low Bit {KV} Cache Quantization with Cross-Layer Compression},
  author    = {Yang, Haoqi  and
               Yao, Yao  and
               Li, Zuchao  and
               Qi, Baoyuan  and
               Guoming, Liu  and
               Zhao, Hai},
  editor    = {Christodoulopoulos, Christos  and
               Chakraborty, Tanmoy  and
               Rose, Carolyn  and
               Peng, Violet},
  booktitle = {Proceedings of the 2025 Conference on Empirical Methods in Natural Language Processing},
  month     = nov,
  year      = {2025},
  address   = {Suzhou, China},
  publisher = {Association for Computational Linguistics},
  url       = {https://aclanthology.org/2025.emnlp-main.494/},
  doi       = {10.18653/v1/2025.emnlp-main.494},
  pages     = {9785--9800},
  isbn      = {979-8-89176-332-6}
}

@inproceedings{kurtic-etal-2025-give,
  title     = {``Give Me {BF}16 or Give Me Death''? Accuracy-Performance Trade-Offs in {LLM} Quantization},
  author    = {Kurtic, Eldar  and
               Marques, Alexandre Noll  and
               Pandit, Shubhra  and
               Kurtz, Mark  and
               Alistarh, Dan},
  editor    = {Che, Wanxiang  and
               Nabende, Joyce  and
               Shutova, Ekaterina  and
               Pilehvar, Mohammad Taher},
  booktitle = {Proceedings of the 63rd Annual Meeting of the Association for Computational Linguistics (Volume 1: Long Papers)},
  month     = jul,
  year      = {2025},
  address   = {Vienna, Austria},
  publisher = {Association for Computational Linguistics},
  url       = {https://aclanthology.org/2025.acl-long.1304/},
  doi       = {10.18653/v1/2025.acl-long.1304},
  pages     = {26872--26886},
  isbn      = {979-8-89176-251-0}
}

@software{Thakkar_CUTLASS_2023,
  author  = {Thakkar, Vijay and Ramani, Pradeep and Cecka, Cris and Shivam, Aniket and Lu, Honghao and Yan, Ethan and Kosaian, Jack and Hoemmen, Mark and Wu, Haicheng and Kerr, Andrew and Nicely, Matt and Merrill, Duane and Blasig, Dustyn and Atluri, Aditya and Qiao, Fengqi and Majcher, Piotr and Springer, Paul and Hohnerbach, Markus and Wang, Jin and Gupta, Manish},
  license = {BSD-3-Clause},
  month   = jan,
  title   = {{CUTLASS}},
  url     = {https://github.com/NVIDIA/cutlass},
  version = {3.0.0},
  year    = {2023}
}

@inproceedings{ashkboos2024slicegpt,
  title     = {Slice{GPT}: Compress Large Language Models by Deleting Rows and Columns},
  author    = {Saleh Ashkboos and Maximilian L. Croci and Marcelo Gennari do Nascimento and Torsten Hoefler and James Hensman},
  booktitle = {The Twelfth International Conference on Learning Representations},
  year      = {2024},
  url       = {https://openreview.net/forum?id=vXxardq6db}
}

@inproceedings{savkin2025nestquant,
  title     = {NestQuant: nested lattice quantization for matrix products and {LLM}s},
  author    = {Semyon Savkin and Eitan Porat and Or Ordentlich and Yury Polyanskiy},
  booktitle = {Forty-second International Conference on Machine Learning},
  year      = {2025},
  url       = {https://openreview.net/forum?id=4OWGON33HE}
}

@inproceedings{xiang2025dfrot,
  title     = {{DFR}ot: Achieving Outlier-Free and Massive Activation-Free for Rotated {LLM}s with Refined Rotation},
  author    = {Jingyang Xiang and Sai Qian Zhang},
  booktitle = {Second Conference on Language Modeling},
  year      = {2025},
  url       = {https://openreview.net/forum?id=WzGypILLDb}
}

@article{krishnamoorthi2018quantizing,
  title   = {Quantizing deep convolutional networks for efficient inference: A whitepaper},
  author  = {Krishnamoorthi, Raghuraman},
  journal = {arXiv preprint arXiv:1806.08342},
  year    = {2018}
}

@inproceedings{
zhang2026qronos,
title={Qronos: Correcting the Past by Shaping the Future... in Post-Training Quantization},
author={Shihao Zhang and Haoyu Zhang and Ian Colbert and Rayan Saab},
booktitle={The Fourteenth International Conference on Learning Representations},
year={2026},
url={https://openreview.net/forum?id=7axclBCYul}
}

@inproceedings{
li2026rethinking,
title={Rethinking Residual Errors in Compensation-based {LLM} Quantization},
author={Shuaiting Li and Juncan Deng and Kedong Xu and Rongtao Deng and Hong Gu and Minghan Jiang and Haibin Shen and Kejie Huang},
booktitle={The Fourteenth International Conference on Learning Representations},
year={2026},
url={https://openreview.net/forum?id=LWYZ1nNkJl}
}

@inproceedings{
darcet2024vision,
title={Vision Transformers Need Registers},
author={Timoth{\'e}e Darcet and Maxime Oquab and Julien Mairal and Piotr Bojanowski},
booktitle={The Twelfth International Conference on Learning Representations},
year={2024},
url={https://openreview.net/forum?id=2dnO3LLiJ1}
}

@inproceedings{wolf-etal-2020-transformers,
    title = "Transformers: State-of-the-Art Natural Language Processing",
    author = "Thomas Wolf and Lysandre Debut and Victor Sanh and Julien Chaumond and Clement Delangue and Anthony Moi and Pierric Cistac and Tim Rault and Rémi Louf and Morgan Funtowicz and Joe Davison and Sam Shleifer and Patrick von Platen and Clara Ma and Yacine Jernite and Julien Plu and Canwen Xu and Teven Le Scao and Sylvain Gugger and Mariama Drame and Quentin Lhoest and Alexander M. Rush",
    booktitle = "Proceedings of the 2020 Conference on Empirical Methods in Natural Language Processing: System Demonstrations",
    month = oct,
    year = "2020",
    address = "Online",
    publisher = "Association for Computational Linguistics",
    url = "https://www.aclweb.org/anthology/2020.emnlp-demos.6",
    pages = "38--45"
}

@article{paszke2017automatic,
  title={Automatic differentiation in pytorch},
  author={Paszke, Adam and Gross, Sam and Chintala, Soumith and Chanan, Gregory and Yang, Edward and DeVito, Zachary and Lin, Zeming and Desmaison, Alban and Antiga, Luca and Lerer, Adam},
  year={2017}
}
}


\appendix

\newpage
\section*{Appendix}\label{app:appendix}

We provide detailed quantization results for Llama and Qwen 2.5 models under W4A4KV4 quantization in Table~\ref{tab:appendix_result_llama} and Table~\ref{tab:appendix_result_qwen}, respectively.

\begin{table}[h]
    \caption{
        Detailed quantization results for Qwen 2.5 models under W4A4KV4 quantization.
        We report the perplexity (PPL \(\downarrow\)) on WikiText~\cite{merity2017pointer} and the 0-shot accuracy across 8 tasks.
        Results for 16-bit, GPTQ~\cite{frantar2023optq}, QUIK~\cite{QUIK}, QuaRot~\cite{ashkboos2024quarot}, and ResQ~\cite{saxena2025resq} are referenced from \cite{saxena2025resq}.
    }\label{tab:appendix_result_qwen}
    \centering
    \setlength{\tabcolsep}{4pt}
    \begin{tabular}{llcccccccccccccc}
        \toprule
                                                       & Method        & \textbf{PPL}    & ARC-c          & ARC-e          & BoolQ          & HellaS         & OBQA           & PIQA           & SIQA           & WinoG          & \textbf{Avg.}  \\
        \midrule
        \multirow{7}{*}{\rotatebox{90}{Qwen 2.5-1.5B}} & 16-bit        & ~~9.3~~         & 45.1           & 72.1           & 72.9           & 67.7           & 40.2           & 76.3           & 48.8           & 63.7           & 60.85          \\
        \cdashline{2-15}\noalign{\vskip 3.5pt}
                                                       & GPTQ          & 25770           & 23.9           & 26.9           & 43.9           & 26.1           & 27.6           & 49.7           & 32.1           & 51.5           & 35.21          \\
                                                       & QUIK          & 6614            & 21.8           & 31.9           & 40.9           & 27.9           & 27.4           & 52.8           & 35.2           & 48.6           & 35.81          \\
                                                       & QuaRot        & 6600            & 23.6           & 37.3           & 46.2           & 28.6           & 27.0           & 56.3           & 35.2           & 52.4           & 38.33          \\
                                                       & ResQ          & 12.5~~          & 38.7           & 64.1           & 65.7           & 61.4           & \textbf{37.8}  & 71.6           & 42.7           & \textbf{60.1}  & 55.26          \\
                                                       & \textbf{OffQ} & \textbf{11.35}  & \textbf{40.87} & \textbf{69.74} & \textbf{70.73} & \textbf{63.27} & \textbf{37.80} & \textbf{73.12} & \textbf{44.93} & 59.75          & \textbf{57.53} \\
        \midrule\addlinespace[-0.1ex]
        \midrule
        \multirow{7}{*}{\rotatebox{90}{Qwen 2.5-3B}}   & 16-bit        & ~~8.0~~         & 47.4           & 73.0           & 77.5           & 73.6           & 42.0           & 78.7           & 49.9           & 68.4           & 63.81          \\
        \cdashline{2-15}\noalign{\vskip 3.5pt}
                                                       & GPTQ          & 9978            & 26.0           & 26.7           & 41.5           & 26.7           & 28.2           & 51.5           & 31.9           & 48.3           & 35.10          \\
                                                       & QUIK          & 15.5~~          & 36.1           & 55.4           & 61.4           & 57.2           & 36.2           & 67.1           & 40.8           & 55.3           & 51.19          \\
                                                       & QuaRot        & 68.8~~          & 32.4           & 53.1           & 51.6           & 49.2           & 33.4           & 66.7           & 39.3           & 56.4           & 47.76          \\
                                                       & ResQ          & ~~9.0~~         & 45.3           & 70.5           & \textbf{72.7}  & \textbf{70.2}  & \textbf{42.4}  & 76.8           & 46.7           & \textbf{64.4}  & 61.13          \\
                                                       & \textbf{OffQ} & ~~\textbf{8.98} & \textbf{47.10} & \textbf{73.90} & 70.76          & 69.94          & 40.60          & \textbf{77.15} & \textbf{48.31} & 64.01          & \textbf{61.47} \\
        \midrule\addlinespace[-0.1ex]
        \midrule
        \multirow{7}{*}{\rotatebox{90}{Qwen 2.5-7B}}   & 16-bit        & ~~6.8~~         & 51.2           & 77.6           & 84.7           & 78.9           & 47.2           & 80.0           & 54.8           & 73.2           & 68.45          \\
        \cdashline{2-15}\noalign{\vskip 3.5pt}
                                                       & GPTQ          & 13594           & 25.2           & 25.6           & 37.8           & 26.3           & 28.2           & 52.4           & 34.4           & 48.9           & 34.85          \\
                                                       & QUIK          & 260.3           & 29.5           & 42.4           & 51.7           & 36.3           & 28.2           & 59.6           & 34.5           & 49.6           & 41.48          \\
                                                       & QuaRot        & 4036            & 25.9           & 41.0           & 39.1           & 29.1           & 27.6           & 57.9           & 35.7           & 50.6           & 38.36          \\
                                                       & ResQ          & ~~8.2~~         & 49.0           & 74.7           & 81.4           & 75.7           & \textbf{45.0}  & \textbf{78.9}  & 49.4           & 68.2           & 65.29          \\
                                                       & \textbf{OffQ} & ~~\textbf{7.66} & \textbf{49.23} & \textbf{74.96} & \textbf{83.30} & \textbf{76.49} & 44.80          & 78.56          & \textbf{52.71} & \textbf{69.22} & \textbf{66.16} \\
        \midrule\addlinespace[-0.1ex]
        \midrule
        \multirow{7}{*}{\rotatebox{90}{Qwen 2.5-14B}}  & 16-bit        & ~~5.3~~         & 58.8           & 79.4           & 85.4           & 82.9           & 45.4           & 81.9           & 55.3           & 75.8           & 70.61          \\
        \cdashline{2-15}\noalign{\vskip 3.5pt}
                                                       & GPTQ          & 5100            & 23.8           & 29.1           & 47.7           & 30.1           & 27.6           & 51.3           & 34.6           & 51.2           & 36.93          \\
                                                       & QUIK          & 10.5~~          & 45.0           & 67.1           & 64.7           & 68.9           & 37.6           & 74.8           & 43.9           & 59.3           & 57.66          \\
                                                       & QuaRot        & ~~6.8~~         & 54.8           & 79.6           & 79.9           & 78.7           & 44.0           & 79.5           & 49.9           & 70.7           & 67.14          \\
                                                       & ResQ          & ~~6.2~~         & \textbf{57.6}  & \textbf{82.1}  & 84.9           & \textbf{81.1}  & \textbf{44.8}  & 80.5           & 51.7           & 70.6           & 69.16          \\
                                                       & \textbf{OffQ} & ~~\textbf{6.07} & 56.40          & 80.26          & \textbf{85.17} & 81.00          & 43.60          & \textbf{80.79} & \textbf{52.46} & \textbf{73.95} & \textbf{69.20} \\
        \midrule\addlinespace[-0.1ex]
        \midrule
        \multirow{7}{*}{\rotatebox{90}{Qwen 2.5-32B}}  & 16-bit        & 5.0             & 55.7           & 78.0           & 87.4           & 84.1           & 44.4           & 82.3           & 56.4           & 75.2           & 70.44          \\
        \cdashline{2-15}\noalign{\vskip 3.5pt}
                                                       & GPTQ          & 3891            & 25.4           & 35.4           & 48.5           & 31.8           & 27.0           & 53.8           & 35.8           & 50.5           & 38.53          \\
                                                       & QUIK          & 9.6             & 41.0           & 64.6           & 74.9           & 72.0           & 39.6           & 75.8           & 44.5           & 60.2           & 59.08          \\
                                                       & QuaRot        & 6.1             & 54.5           & 76.1           & 85.1           & 81.5           & 44.2           & 80.1           & 51.3           & 70.4           & 67.90          \\
                                                       & ResQ          & ~~5.6~~         & 55.1           & \textbf{78.4}  & 86.0           & 82.5           & \textbf{45.4}  & 81.1           & \textbf{53.9}  & 74.0           & 69.55          \\
                                                       & \textbf{OffQ} & ~~\textbf{5.52} & \textbf{55.12} & 78.07          & \textbf{87.03} & \textbf{82.58} & 45.00          & \textbf{81.94} & 52.05          & \textbf{74.90} & \textbf{69.59} \\
        \midrule\addlinespace[-0.1ex]
        \midrule
        \multirow{7}{*}{\rotatebox{90}{Qwen 2.5-72B}}  & 16-bit        & ~~3.9~~         & 62.6           & 83.2           & 89.2           & 86.0           & 46.6           & 83.6           & 58.4           & 77.7           & 73.41           \\
        \cdashline{2-15}\noalign{\vskip 3.5pt}
                                                       & GPTQ          & 37967           & 25.4           & 25.8           & 38.1           & 25.6           & 26.6           & 51.2           & 34.2           & 49.4           & 34.54           \\
                                                       & QUIK          & 8.3             & 45.1           & 68.1           & 77.2           & 77.2           & 39.0           & 77.4           & 45.6           & 65.6           & 61.90           \\
                                                       & QuaRot        & 4.9             & 55.8           & 81.1           & 87.5           & 84.0           & 45.2           & 81.7           & 52.5           & 74.5           & 70.28           \\
                                                       & ResQ          & ~~4.6~~         & 58.4           & 80.9           & 88.4           & 84.9           & \textbf{48.2}  & 82.6           & 55.5           & \textbf{77.0}  & 71.98           \\
                                                       & \textbf{OffQ} & ~~\textbf{4.29} & \textbf{61.95} & \textbf{81.99} & \textbf{88.93} & \textbf{85.28} & 46.80          & \textbf{83.46} & \textbf{56.50} & 76.56          & \textbf{72.68} \\
        \bottomrule
    \end{tabular}
\end{table}

\begin{table}[t]
    \caption{
        Detailed quantization results for Llama models under W4A4KV4 quantization.
        We report the perplexity (PPL \(\downarrow\)) on WikiText~\cite{merity2017pointer} and the 0-shot accuracy across 8 tasks.
        Results for 16-bit, GPTQ~\cite{frantar2023optq}, QUIK~\cite{QUIK}, QuaRot~\cite{ashkboos2024quarot}, SpinQuant~\cite{liu2025spinquant}, and ResQ~\cite{saxena2025resq} are referenced from \cite{saxena2025resq};
        and results for DFRot~\cite{xiang2025dfrot}, KurTail~\cite{akhondzadeh-etal-2025-kurtail}, and OSTQuant~\cite{hu2025ostquant} are referenced from their respective papers.
    }\label{tab:appendix_result_llama}
    \centering
    \setlength{\tabcolsep}{4.6pt}
    \small
    \renewcommand{\arraystretch}{0.95}
    \begin{tabular}{llcccccccccccccc}
        \toprule
                                                       & Method        & \textbf{PPL}     & ARC-c          & ARC-e          & BoolQ          & HellaS         & OBQA           & PIQA           & SIQA           & WinoG          & \textbf{Avg.}  \\
        \midrule
        \multirow{11}{*}{\rotatebox{90}{Llama 3-8B}}   & 16-bit        & ~~6.1            & 53.2           & 77.1           & 81.1           & 79.2           & 44.8           & 80.9           & 47.0           & 73.4           & 67.09          \\
        \cdashline{2-15}\noalign{\vskip 2pt}
                                                       & GPTQ          & 166.3            & 24.7           & 37.7           & 44.3           & 36.8           & 27.0           & 57.6           & 36.4           & 53.8           & 39.79          \\
                                                       & QUIK          & 14.2~~           & 33.6           & 56.4           & 60.5           & 61.5           & 33.2           & 68.7           & 39.9           & 59.0           & 51.60          \\
                                                       & QuaRot        & ~~7.8~~          & 45.1           & 70.4           & 73.8           & 74.7           & 42.6           & 76.6           & 45.1           & 68.5           & 62.10          \\
                                                       & SpinQuant     & ~~7.4~~          & 48.0           & 75.4           & 75.8           & 75.4           & \textbf{43.8}  & 77.5           & 45.0           & 69.2           & 63.76          \\
                                                       & DFRot         & ~~7.91           & 44.97          & 71.09          & 73.27          & 74.13          & 43.00          & 78.24          & 44.58          & 69.53          & 62.35          \\
                                                       & KurTail       & ~~7.2~~          & 48.2           & 75.4           & 79.2           & 76.4           & 43.6           & 78.4           & 45.8           & 70.0           & 64.63          \\
                                                       & OSTQuant      & ~~7.29           & 49.32          & 76.73          & 78.87          & 76.01          & 43.20          & 78.51          & 45.70          & 69.22          & 64.70          \\
                                                       & ResQ          & ~~7.1~~          & 49.2           & 75.0           & 72.5           & 76.5           & 43.0           & 78.3           & 45.8           & \textbf{71.0}  & 63.91          \\
                                                       & \textbf{OffQ} & ~~\textbf{6.98}  & \textbf{50.68} & \textbf{77.44} & \textbf{80.43} & \textbf{76.96} & \textbf{43.80} & \textbf{78.89} & \textbf{45.96} & 69.77          & \textbf{65.49} \\
        \midrule\addlinespace[-0.1ex]
        \midrule
        \multirow{11}{*}{\rotatebox{90}{Llama 3-70B}}  & 16-bit        & ~~ 2.9           & 64.2           & 85.9           & 85.3           & 84.9           & 48.6           & 84.4           & 50.8           & 80.6           & 73.09          \\
        \cdashline{2-15}\noalign{\vskip 2pt}
                                                       & GPTQ          & 11655.0          & 25.9           & 26.0           & 37.9           & 26.2           & 28.6           & 50.4           & 34.3           & 49.9           & 34.90          \\
                                                       & QUIK          & ~~8.0~~          & 44.5           & 68.9           & 60.7           & 75.0           & 36.4           & 76.1           & 43.2           & 60.4           & 58.15          \\
                                                       & QuaRot        & ~~5.7~~          & 53.7           & 74.5           & 81.6           & 81.1           & 46.6           & 81.0           & 46.8           & 75.2           & 67.56          \\
                                                       & SpinQuant     & ~~6.2~~          & 52.0           & 77.3           & 81.7           & 75.6           & 43.8           & 78.8           & 43.4           & 72.8           & 65.68          \\
                                                       & DFRot         & ~~5.03           & 58.02          & 81.1           & 81.13          & 81.59          & 47.4           & 81.83          & 46.57          & 74.19          & 68.98          \\
                                                       & KurTail       & ~~4.2~~          & 59.2           & 82.7           & 83.9           & 83.3           & 46.6           & \textbf{83.5}  & 49.7           & 76.6           & 70.69          \\
                                                       & OSTQuant      & ~~4.01           & 61.29          & 82.39          & 83.43          & 83.25          & \textbf{48.93} & 81.73          & \textbf{51.24} & 77.01          & \textbf{71.16} \\
                                                       & ResQ          & ~~4.1~~          & \textbf{61.4}  & \textbf{84.3}  & 83.9           & 83.5           & 46.0           & 83.1           & 48.6           & \textbf{78.3}  & 71.14          \\
                                                       & \textbf{OffQ} & ~~\textbf{3.88}  & 59.22          & 79.12          & \textbf{86.12} & \textbf{84.21} & 46.80          & 82.43          & 48.82          & \textbf{78.30} & 70.63          \\
        \midrule\addlinespace[-0.1ex]
        \midrule
        \multirow{11}{*}{\rotatebox{90}{Llama 3-2-1B}} & 16-bit        & ~~9.8~~          & 36.5           & 60.6           & 63.4           & 63.6           & 37.4           & 74.5           & 42.8           & 60.1           & 54.86          \\
        \cdashline{2-15}\noalign{\vskip 2pt}
                                                       & GPTQ          & 108.9            & 24.7           & 32.7           & 52.3           & 30.7           & 23.6           & 54.3           & 34.4           & 51.1           & 37.98          \\
                                                       & QUIK          & 21.8~~           & 27.4           & 46.0           & 55.0           & 46.0           & 26.4           & 62.4           & 38.6           & 52.6           & 44.30          \\
                                                       & QuaRot        & 14.3~~           & 30.0           & 51.4           & 59.1           & 54.0           & 34.2           & 66.7           & 39.6           & \textbf{57.1}  & 49.01          \\
                                                       & SpinQuant     & 13.6~~           & 32.3           & 51.8           & 59.3           & 55.4           & 30.4           & 67.7           & 38.6           & 54.7           & 48.78          \\
                                                       & KurTail       & 12.9~~           & 31.1           & 52.9           & 60.7           & 56.4           & \textbf{36.4}  & 68.6           & 40.5           & 54.3           & 50.11          \\
                                                       & ResQ          & 12.4~~           & \textbf{34.0}  & 54.2           & 57.0           & 57.3           & 31.2           & 69.4           & \textbf{41.0}  & 56.8           & 50.11          \\
                                                       & \textbf{OffQ} & \textbf{12.32}   & 30.80          & \textbf{55.22} & \textbf{62.26} & \textbf{57.55} & 33.2           & \textbf{70.95} & 40.89          & 56.43          & \textbf{50.91} \\
        \midrule\addlinespace[-0.1ex]
        \midrule
        \multirow{11}{*}{\rotatebox{90}{Llama 3-2-3B}} & 16-bit        & ~~7.8~~          & 46.2           & 71.7           & 73.1           & 73.7           & 43.4           & 77.4           & 47.2           & 69.1           & 62.73          \\
        \cdashline{2-15}\noalign{\vskip 2pt}
                                                       & GPTQ          & 178.3            & 27.0           & 27.0           & 48.8           & 44.4           & 27.8           & 59.1           & 37.1           & 51.5           & 40.34          \\
                                                       & QUIK          & 15.8~~           & 32.9           & 50.1           & 52.6           & 59.1           & 33.2           & 68.7           & 40.3           & 53.0           & 48.74          \\
                                                       & QuaRot        & 10.1~~           & 38.6           & 59.0           & 65.9           & 66.5           & 35.8           & 74.4           & 43.1           & 65.2           & 56.06          \\
                                                       & SpinQuant     & ~~9.2~~          & 38.9           & 64.8           & 68.0           & 69.1           & 39.4           & 74.9           & 45.1           & 62.9           & 57.89          \\
                                                       & KurTail       & ~~9.0~~          & 42.2           & 66.7           & 69.8           & 68.8           & \textbf{39.8}  & \textbf{75.6}  & 44.8           & 64.6           & 59.04          \\
                                                       & ResQ          & ~~8.8~~          & 43.1           & 65.6           & 68.8           & 70.5           & 38.4           & 75.1           & 45.6           & 64.8           & 58.99          \\
                                                       & \textbf{OffQ} & ~~\textbf{8.78}  & \textbf{44.8}  & \textbf{70.41} & \textbf{71.62} & \textbf{71.49} & 39.60          & 75.52          & \textbf{45.65} & \textbf{67.32} & \textbf{60.80} \\
        \midrule\addlinespace[-0.1ex]
        \midrule
        \multirow{11}{*}{\rotatebox{90}{Llama 2-7B}}   & 16-bit        & ~~5.5~~          & 46.3           & 74.6           & 77.8           & 75.9           & 44.2           & 79.2           & 46.1           & 69.1           & 64.15          \\
        \cdashline{2-15}\noalign{\vskip 2pt}
                                                       & GPTQ          & 9600             & 24.8           & 31.4           & 55.4           & 30.6           & 25.6           & 55.8           & 34.2           & 53.3           & 38.89          \\
                                                       & QUIK          & ~~7.5~~          & 39.8           & 63.7           & 68.9           & 68.3           & 37.8           & 72.9           & 42.1           & 62.4           & 56.99          \\
                                                       & QuaRot        & ~~6.1~~          & 41.5           & 71.4           & 73.2           & 73.2           & 40.6           & 76.9           & 43.6           & 65.6           & 60.75          \\
                                                       & SpinQuant     & ~~6.0~~          & 43.6           & 71.3           & 73.8           & 73.2           & 40.4           & 76.0           & 44.1           & 65.4           & 60.98          \\
                                                       & DFRot         & ~~6.25           & 43.52          & 70.83          & 73.3           & 72.62          & 41.40          & 76.82          & 44.17          & 65.11          & 60.97          \\
                                                       & KurTail       & ~~5.9~~          & 43.1           & 72.0           & 72.0           & 73.2           & 41.2           & 76.6           & 45.6           & 66.8           & 61.31          \\
                                                       & OSTQuant      & ~~5.91           & 42.92          & 72.56          & 74.71          & 73.14          & \textbf{44.40} & 77.42          & \textbf{44.98} & 66.77          & \textbf{62.11} \\
                                                       & ResQ          & ~~5.8~~          & \textbf{44.0}  & \textbf{72.6}  & \textbf{75.3}  & \textbf{74.0}  & 41.0           & 77.9           & 43.9           & 66.9           & 61.95          \\
                                                       & \textbf{OffQ} & ~~\textbf{5.77}  & 43.77          & 70.79          & 74.89          & 73.84          & 41.80          & \textbf{78.02} & 44.27          & \textbf{68.51} & 61.99          \\
        \midrule\addlinespace[-0.1ex]
        \midrule
        \multirow{11}{*}{\rotatebox{90}{Llama 2-13B}}  & 16-bit        & ~~4.9~~          & 49.1           & 77.4           & 80.5           & 79.4           & 45.2           & 80.7           & 47.2           & 72.1           & 66.45          \\
        \cdashline{2-15}\noalign{\vskip 2pt}
                                                       & GPTQ          & 3120             & 23.6           & 31.1           & 38.7           & 27.2           & 26.8           & 53.6           & 35.8           & 49.8           & 35.83          \\
                                                       & QUIK          & ~~6.8~~          & 43.7           & 68.0           & 71.3           & 73.3           & 40.0           & 75.7           & 45.1           & 64.6           & 60.21          \\
                                                       & QuaRot        & ~~5.4~~          & 46.9           & 74.9           & 76.6           & 75.8           & 42.6           & 79.1           & 45.5           & 69.0           & 63.80          \\
                                                       & SpinQuant     & ~~5.2~~          & 49.0           & \textbf{76.3}  & 78.2           & 77.1           & 42.8           & \textbf{79.3}  & 46.3           & 69.5           & 64.81          \\
                                                       & DFRot         & ~~5.43           & 46.50          & 73.48          & 76.67          & 76.83          & 43.00          & 79.27          & 45.55          & 69.30          & 63.83          \\
                                                       & KurTail       & ~~5.2~~          & 48.1           & 75.4           & 79.7           & 77.4           & \textbf{45.0}  & 79.0           & 45.6           & \textbf{71.2}  & 65.18          \\
                                                       & OSTQuant      & ~~5.25           & 47.10          & 75.21          & 77.46          & 76.71          & 44.60          & 78.67          & 45.75          & 68.03          & 64.19          \\
                                                       & ResQ          & ~~\textbf{5.1}~~ & \textbf{49.1}  & 76.1           & \textbf{79.7}  & 77.9           & 43.6           & 79.1           & \textbf{46.6}  & 69.9           & \textbf{65.25} \\
                                                       & \textbf{OffQ} & ~~5.11           & 48.38          & \textbf{76.30} & 79.36          & \textbf{77.96} & 43.6           & 79.05          & 46.52          & 70.8           & \textbf{65.25} \\
        \bottomrule
    \end{tabular}
\end{table}


\end{document}